\documentclass{article} 
\usepackage{collas2022_conference,times}

\usepackage[utf8]{inputenc} 
\usepackage[T1]{fontenc}    
\usepackage{hyperref}       
\usepackage{url}            
\usepackage{booktabs}       
\usepackage{amsfonts}       
\usepackage{nicefrac}       
\usepackage{microtype}      
\usepackage{xcolor}         
\usepackage{graphicx}
\usepackage{subfigure}
\usepackage{wrapfig}
\usepackage{soul}
\usepackage{ulem}
\usepackage{wrapfig}

\usepackage{algorithm}
\usepackage[noend]{algpseudocode}

\usepackage{multirow}
\usepackage{arydshln}
\makeatletter
\def\adl@drawiv#1#2#3{%
        \hskip.5\tabcolsep
        \xleaders#3{#2.5\@tempdimb #1{1}#2.5\@tempdimb}%
                #2\z@ plus1fil minus1fil\relax
        \hskip.5\tabcolsep}
\newcommand{\cdashlinelr}[1]{%
  \noalign{\vskip\aboverulesep
           \global\let\@dashdrawstore\adl@draw
           \global\let\adl@draw\adl@drawiv}
  \cdashline{#1}
  \noalign{\global\let\adl@draw\@dashdrawstore
           \vskip\belowrulesep}}
\makeatother
\newcommand\blfootnote[1]{%
  \begingroup
  \renewcommand\thefootnote{}\footnote{#1}%
  \addtocounter{footnote}{-1}%
  \endgroup
}

\newcommand{\Single}{\textit{SM}}
\newcommand{\Ens}{\textit{Ens}}
\newcommand{\UMix}{\textit{UMix}}
\newcommand{\FF}{\textit{FF}}
\newcommand{\MoE}{\textit{MoE}}
\newcommand{\gMoE}{\textit{gMoE}}
\newcommand{\gEns}{\textit{gEns}}

\newcommand{\mr}[1]{\textcolor{purple}{MR: #1}}

\newcommand{\comment}[1]{}

\newcommand{\OO}[1]{\textcolor{orange}{#1}}
\newcommand{\BB}[1]{\textcolor{blue}{#1}}
\newcommand{\NEW}[1]{\textcolor{black}{#1}}

\usepackage{amsmath,amsfonts,bm}

\def\eqref#1{equation~\ref{#1}}

\def\1{\bm{1}}

\DeclareMathAlphabet{\mathsfit}{\encodingdefault}{\sfdefault}{m}{sl}
\SetMathAlphabet{\mathsfit}{bold}{\encodingdefault}{\sfdefault}{bx}{n}

\usepackage{hyperref}
\usepackage{url}

\title{On Anytime Learning at Macroscale}

\author{Lucas Caccia
~~~~~~~~~~~~   \\
McGill University, Mila \\
Facebook AI Research \\
\And
Jing Xu  ~~~~~~~~~~~~~~~~~ \\
Facebook AI Research 
\And 
Myle Ott ~~~~~~~~~~~~~~~~ \\
Facebook AI Research
\AND
~~~~~~~~~~~~~~~~ ~~~~~~~~~~~~~~~~Marc'Aurelio Ranzato$^{\dagger}$$^\ast$ 
~~~~~~~~~~~~~~~~\\
~~~~~~~~~~~~~~~~ ~~~~~~~~~~~~~~~~Facebook AI Research
\And
Ludovic Denoyer$^{\ddagger}$$^\ast$
~~~~~~~~~~~~~~~~~~~~~~~~~  \\
Facebook AI Research 
}

\collasfinalcopy 
\begin{document}

\maketitle

\begin{abstract}
In many practical applications of machine learning data arrives sequentially over time in large chunks. Practitioners have then to decide how to allocate their computational budget in order to obtain the best performance at any point in time. Online learning theory for convex optimization suggests that the best strategy is to use data as soon as it arrives. However, this might not be the best strategy when using deep non-linear networks, particularly when these perform multiple passes over each chunk of data rendering the overall distribution non i.i.d.. In this paper, we formalize this learning setting in the simplest scenario in which each data chunk is drawn from the same underlying distribution, and make a first attempt at empirically answering the following questions: How long should the learner wait before training on the newly arrived chunks? What architecture should the learner adopt? Should the learner increase capacity over time as more data is observed? We probe this learning setting using convolutional neural networks trained on classic computer vision benchmarks as well as a large transformer model trained on a large-scale language modeling task. Code is available at \url{www.github.com/facebookresearch/ALMA}.

      \blfootnote  {* Authors contributed equally} \\
      \blfootnote{  $\dagger$ Now at DeepMind} \\
       \blfootnote{ $\ddagger$ Now at Ubisoft}

\end{abstract}

\section{Introduction}
In many practical applications of machine learning, data is not static but arrives sequentially in large chunks (or \textcolor{black}{mega-}batches). For instance, deployed language modeling systems need to be updated every few months to accommodate new snapshots of the Common Crawl dataset\footnote{\url{https://commoncrawl.org/the-data/}}. Similarly, visual object recognition systems need to be updated as new labeled data is gathered thanks to users interacting with the system. Moreover, as computing clusters are equipped with more memory and compute,  machine learning practitioners would like to train bigger and bigger models on the ever increasing amount of data, since bigger models are often more accurate. In this setting, they face a dilemma: How to maximize the performance of the system at any point in time while satisfying a certain computational budget?

This question has certainly been studied before, most notably in the online learning literature~\citep{cesa2006prediction}. For instance, in a contextual bandit setting the learner observes one example at the time and receives a reward after making a prediction. Of course, this can be extended to the case where the input is not just a single example but a set of examples (hereinafter referred to as \textcolor{black}{mega-batch}). 


While prior works on online learning set a sound theoretical framework, there are some subtle issues that make it not quite applicable to the practical setting described above. First, computation is seldom explicitly taken into account, while in practice algorithms that are too computationally intensive cannot be considered at scale. Second, the vast majority of these works assumes linearity of predictors and convexity of optimization problems, whereby the order of examples does not change the optimum solution. Instead, in many practical applications (like language modeling) we are interested in using deep neural networks which are highly non-linear and which map to non-convex optimization problems. The lack of linearity hinders theoretical analysis, and it has profound practical implications. For instance, according to online learning theory the best case scenario is achieved when there are no delays~\citep{Joulani16, Flaspohler21}, meaning that examples and their error signal are best to be consumed right away without any staleness in the model parameters. To use the language of the practitioner training a language model, this means that according to convex online learning the best strategy is to train one \textcolor{black}{mega-batch} at the time. This however might not be a good strategy.

\begin{figure}[t]
\vspace{-30pt}
  \centering
\includegraphics[width=.45\textwidth]{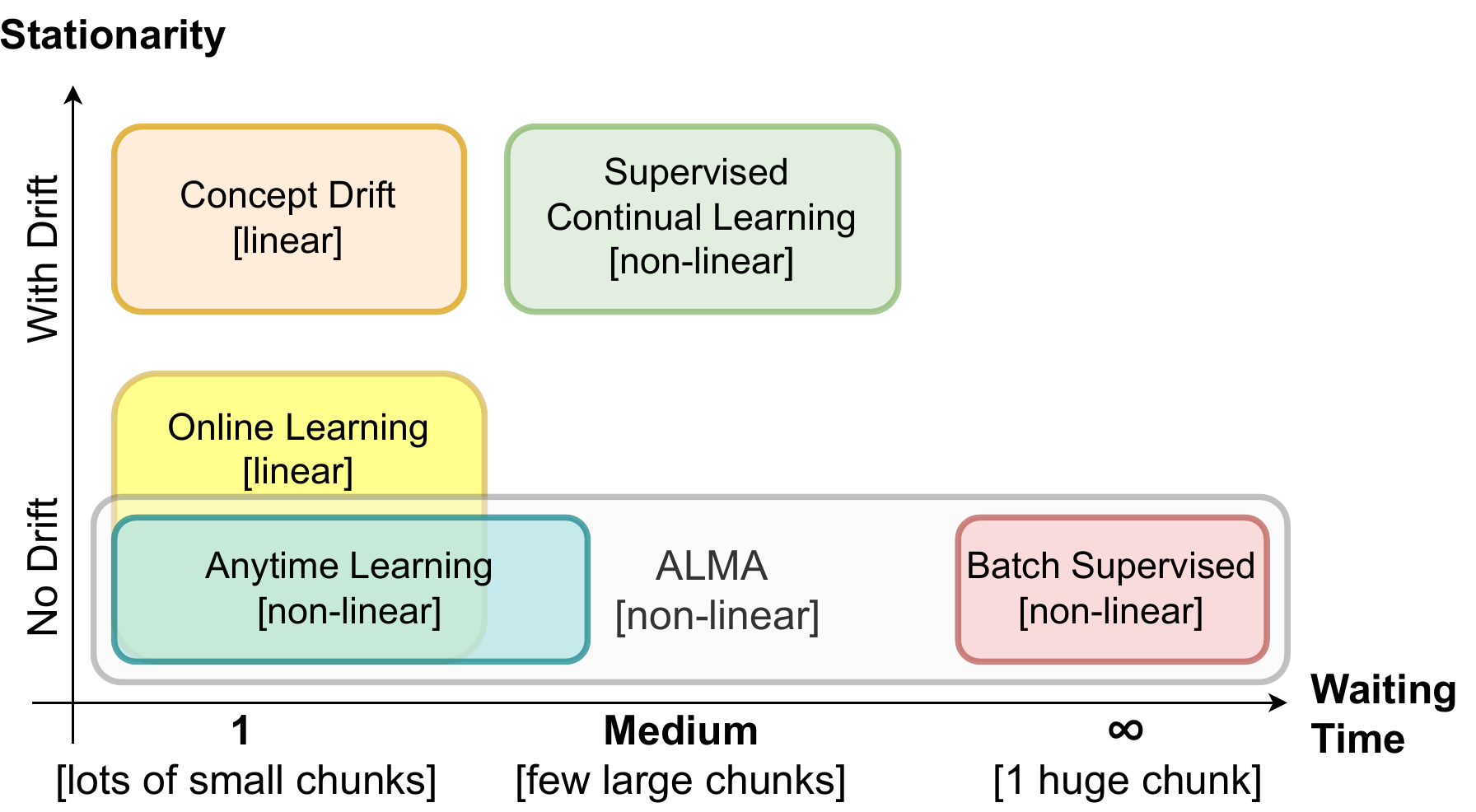}
\includegraphics[width=.45\textwidth]{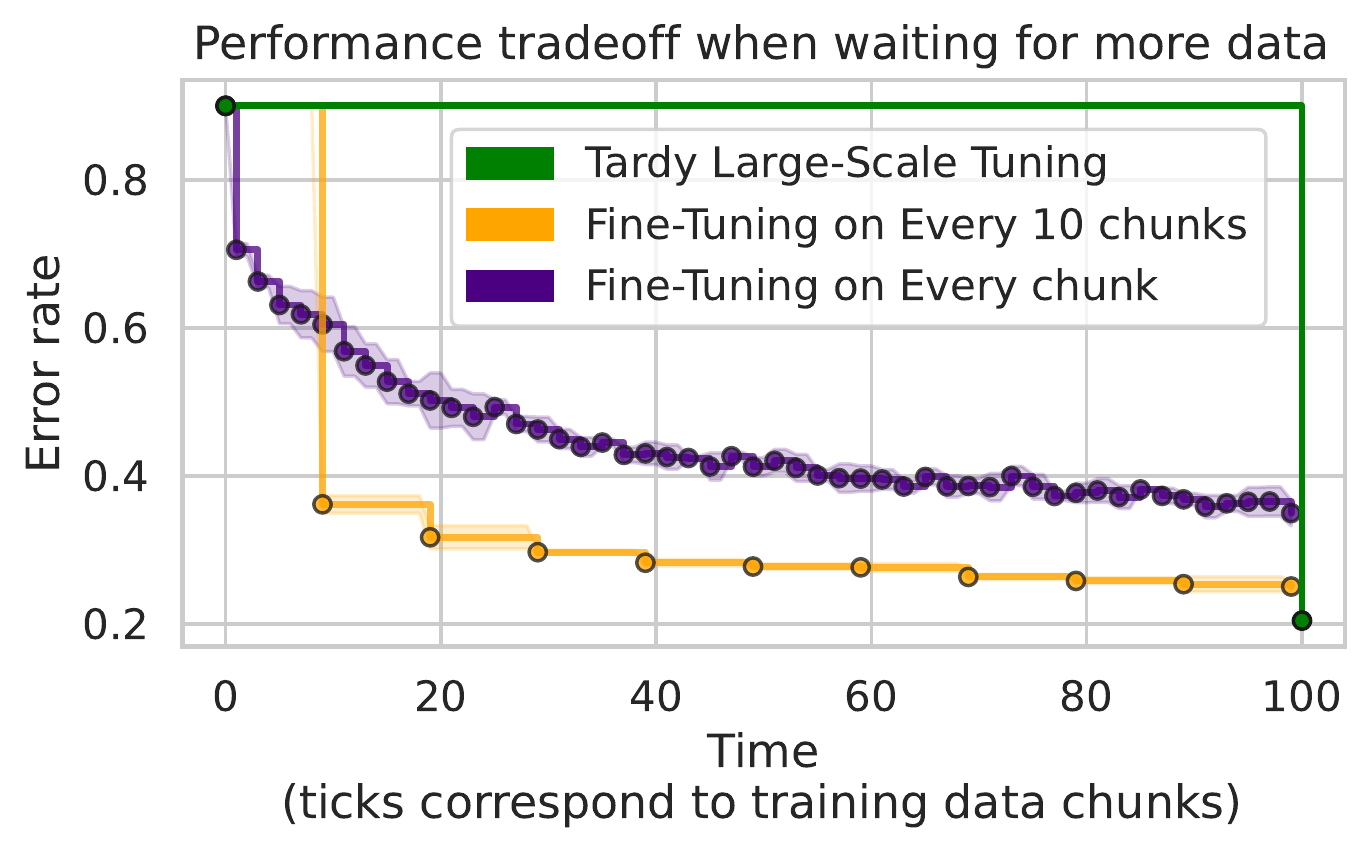}
  \caption{\small Left: ALMA compared to other learning frameworks. In ALMA, \textcolor{black}{mega-batches} of data are drawn from the same distribution (no drift) and arrive sequentially, but the learner can decide how long to wait before training on them. In the limit, if the learner waits till the end of the stream then learning reduces to standard batch supervised learning. Right: Examples of CIFAR~10 learning curves varying how long to wait before updating the model. Waiting for a small number of \textcolor{black}{mega-batches} before updating the parameters results in lower anytime error rate (smaller area under the learning curve).}  \label{fig:alma_context}
\end{figure}

Consider what would happen if the deep neural network does multiple passes over each \textcolor{black}{mega-batch} before processing the next, and compare its performance to the one of a learner that waits for all the \textcolor{black}{mega-batches} to arrive before shuffling all data and applying the same stochastic gradient descent optimization algorithm as shown on the right part of Fig.~\ref{fig:alma_context}.  The latter setting is the standard procedure used in supervised learning (green curve): the learning algorithm optimizes a fixed objective (i.e the empirical risk over the entire training dataset) that is known to produce good predictors. While this predictor obtains the best final performance, it also attains the worst anytime performance since its predictions were random throughout the learning experience. In the former setting, by updating after each new \textcolor{black}{mega-batch} (purple curve), we can expect to maintain a good predictor all along the training experience, overcoming the problem described previously. However in this case, the learner is facing a changing learning objective, since each new \textcolor{black}{mega-batch} defines a slightly different empirical risk~\citep{Jothimurugesan18}. While we can expect this effect to be negligible when using linear models which eventually will converge to the same global optimum when all \textcolor{black}{mega-batches} are available, this is not the case when using non-linear predictors like deep neural networks. In that case, the sequence of optimization problems generated by the sequence of \textcolor{black}{mega-batches} may lead the learner to a completely different (local) optimum than the supervised learning setting, and thus to a completely different predictor. There is thus an open question about how different models behave when performing sequential learning over a stream of \textcolor{black}{mega-batches}. 


In this paper, we empirically analyze several deep learning models (\textsection\ref{sec:algo}) under the assumption that data comes as a sequence of \textcolor{black}{mega-batches}, all drawn from the same distribution for simplicity. Since we are interested in models that attain good performance at any point in time and since we evaluate only after learning on each \textcolor{black}{mega-batch} but not during the learning of each individual \textcolor{black}{mega-batch}, we dub this learning setting Anytime Learning at MAcroscale (ALMA) (\textsection\ref{sec:setting}).

Through extensive empirical analysis (\textsection\ref{sec:experiments}) we provide supporting evidence that waiting for a few \textcolor{black}{mega-batches} before updating the model is often the best strategy, although how long to wait depends on several factors such as the time horizon and model size relative to the amount of data in each \textcolor{black}{mega-batch}. Second, bigger models are more statistically efficient and generalize better. Third, none of the approaches we tried for growing the architecture were more effective than simpler alternatives which used fixed architectures, like ensembling. Overall, this study provides clear directions of future research, and also a platform for benchmarking new approaches against well tuned baselines (code available in supplementary material).

\section{Related Work} \label{sec:related_work}

ALMA relates to several other learning
frameworks as illustrated on the left of Figure~\ref{fig:alma_context}.
i) It shares the same assumptions of classical batch supervised learning~\citep{erm} at the level of each \textcolor{black}{mega-batch}. However, it overall violates the assumptions of i.i.d.~observations, because data points
come in a stream of \textcolor{black}{mega-batches} and because the learner typically makes several passes over each \textcolor{black}{mega-batch}. Moreover, in ALMA the learner can choose how long to wait before training. In this sense, batch supervised learning can be thought of as an extreme case of ALMA (single \textcolor{black}{mega-batch} because learner waited till the end of the stream to train).
ii) As mentioned in the previous section, ALMA relates to online learning~\citep{bottou-98x} because data comes sequentially and because in both cases we measure performance in terms of regret (although in \textsection\ref{sec:metrics} our cumulative error lacks a reference oracle since this is not known in our setting). However, in ALMA we are also explicit about the computational budget used by the model and aim at striking a good trade-off between regret and computational cost. In our current work, we restrict ALMA to stationary distributions, while online learning is more general and encompasses also non-stationary distributions. Finally and most importantly, in ALMA we focus on non-linear predictors while typical literature on online learning considers linear predictors. iii) Similarly, ALMA relates to concept drift~\citep{lu2018learning} because of the sequential nature of the observations. However, literature on concept drift often focuses on linear predictors. iv) ALMA can be seen as a degenerate case of supervised continual learning, where the task distribution is stationary. However, in supervised continual learning there is often a focus on attaining a predictor that represents the entire distribution of tasks by the end of learning, while in ALMA we measure cumulative error like in prequential learning. v) ALMA relates more broadly to transfer learning~\citep{transfer_learning}, as the problem of adapting to a new batch of data can be interpreted as leveraging knowledge acquired on previous batches to more effciently learn from the new batch of data. 
vi) Finally, ALMA relates to anytime learning~\citep{grefenstette92, ramsey94}, which has been recently applied to compare various autoML frameworks~\citep{pmlrv123liu20a}. \textcolor{black}{However, unlike traditional anytime learning}, in this work we are not interested in assessing the anytime
learning ability at the level of each mega-batch, but only at a coarser granularity, at the level of the entire stream of \textcolor{black}{mega-batches}. \textcolor{black}{Lastly, we note that while anytime learning operates in a similar setting as online learning (see Fig. \ref{fig:alma_context}), it is often used with non-linear predictors in a supervised learning setting.}

To the best of our knowledge, the most relevant prior work is by~\citet{sahoo18} which considers a setting similar to ours, except that their stream is composed by individual examples and in their setting there is no concept of waiting time nor revisiting data points several times. However, they also benchmark against methods that increase capacity over time, although their analysis was limited to fully connected networks.

\section{Learning Setting} \label{sec:setting}
\vspace{-7pt}

In anytime learning at macroscale (ALMA), we assume that
there exists an underlying data distribution $p(x,y)$
with input $x \in \mathbb{R}^D$ and desired label $y \in \{1, \dots, C\}$.
For the sake of simplicity of exposition, in this work we restrict ourselves to classification problems, but similar arguments can be made for regression, for instance. 
The key property of ALMA is that data is presented to the
learner as a stream $\mathcal{S}_B$ of $B$ consecutive batches of examples.
Let $\mathcal{D}_i$ be a collection of $N \gg 0$ i.i.d. samples
randomly drawn from
$p(x,y)$, for $i \in \{1, \dots, B \}$. The stream is
then defined as the ordered sequence $\mathcal{S}_B = \{\mathcal{D}_1,
\dots, \mathcal{D}_B \}$. We refer to each
dataset $\mathcal{D}_i$ as {\em mega-batch}, as it is composed by a
large number of examples. 

Typically a learner $m: \mathbb{R}^D\rightarrow \{1, \dots, C\}$ updates its parameters by processing a
{\em mini-batch} of $n \ll N$ examples at the time from each mega-batch
$\mathcal{D}_i$ in such a way to minimize its objective function. Since the data is observed as a stream of mega-batches, the learner cannot have access to future mega-batches, and cross-validation of model hyper-parameters can only be performed using a subset of the current mega-batch.  In other words, the learner can only do one pass over the stream. However, the learner typically does multiple passes over the current mega-batch if this improves its generalization ability. In fact, the learner might make several passes over the current and some previous mega-batches, although replaying too much might eventually deplete its computational budget.

Either way, since the learner makes several passes over each mega-batch, the overall data distribution observed by the learner \textcolor{black}{by the end of the stream} is not i.i.d., even though mega-batches are drawn from the same underlying distribution $p(x,y)$ and samples drawn from each mega-batch are i.i.d.. \textcolor{black}{For instance, in the limit case where each mega-batch consists of a single example from a set of $n$ examples and a learner performing $k$ passes over each mega-batch, the stream will consist of a sequence of examples (in a certain order) each replicated $k$ times, which is different from  drawing uniformly at random $k*n$ examples from the original set of $n$ examples.} This implies a trade-off between fitting the current data well versus generalizing well by the end of the stream.

In ALMA, the learner has an additional hyper-parameter compared to other learning frameworks: It can decide how long to wait before updating its parameters. We measure such {\em waiting time} in terms of number of consecutive mega-batches. For instance, a model with a waiting time equal to $k$, aggregates $k$ consecutive mega-batches before updating its parameters. This will sacrifice a bit its performance during the waiting period, but might ultimately yield better generalization since the model can better shuffle the data and get closer to the ideal i.i.d. data distribution required by stochastic gradient descent optimization.  

\subsection{Metrics} \label{sec:metrics}
We evaluate learners in the ALMA setting across three axes, namely:
error rate, memory and computation. Let $t$ be the time at which the
$t$-th mega-batch arrives; this data can be used by the model to update its
parameters or it is simply aggregated to previous mega-batches for later use.

We compute the error rate of model $m$ at
time $t$ (after the arrival and potential update over the $t$-th mega-batch) and
compute the area under the curve obtained varying $t$ from $0$ till
the total number of mega-batches $B$; the resulting cumulative error
rate (CER) is: 
\begin{equation}
  \mbox{CER} = \sum_{t=1}^B \frac{1}{|\mathcal{D}^{\mbox{\texttt{\tiny Ts}}} |} \sum_{(x,y) \in
    \mathcal{D}^{\mbox{\texttt{\tiny Ts}}}} |m(x; \theta_t) \neq y| \label{eq:er}
\end{equation}
where $m(x;\theta_t)$ is the model at time $t$ equipped with parameters $\theta_t$,
$\mathcal{D}^{\mbox{\texttt{\tiny Ts}}}$ is the test set (common for all mega-batches in the stream),  $|\mathcal{D}^{\mbox{\texttt{\tiny Ts}}}
|$ is the number of examples in the test set, and $|m(x; \theta_t) \neq y|$
is one if the model prediction does not match the ground truth label
and zero otherwise. The outer sum computes the discrete integral of
the error rate over time. CER is going to be small only when the error rate
is small throughout the entire stream. CER is instead large for a
tardy model that waits till the very last mega-batch to update the
model, even though eventually this may obtain a very low final error
rate.

Similarly, we compute the cumulative memory usage and compute as:
\begin{equation}
  \mbox{Mem} = \sum_{t=0}^B |\theta_t|, \mbox{\hspace{.2cm} }
  \mbox{Comp} = \sum_{t=0}^B \mathcal{O}(m(\cdot; \theta_t)) \label{eq:mem_comp}
\end{equation}
where $|\theta_t|$ is the number of free parameters of the model at
time $t$, and $\mathcal{O}(m(\cdot; \theta_t))$ is the number of flops used by the model to process the $t$-th mega-batch.

Notice that the above metrics are measured by the environment as training progresses, and will be used in our empirical assessment (\textsection\ref{sec:experiments}). However, the learner does not have access to the test set. The learner has only access to the validation set of the current mega-batch, and can only use that to select its own hyper-parameters.

\section{Learning Algorithms} \label{sec:algo}
\vspace{-7pt}

In this section, we describe the methods we tested in the
ALMA setting. They generally follow the learning procedure shown in Algorithm~\ref{algo}.
At a high level, we consider two families of models, those with a monolithic architecture and those with a modular architecture (e.g. ensembling). The latter are amenable to grow over time by adding new modules to the existing set. We will start by describing fixed architectures (\textsection\ref{sec:fixed}) and then conclude with growing architectures (\textsection\ref{sec:grow}). We also evaluate models in the setting where they can replay previous mega-batches.

\begin{algorithm}[t]
  \caption{Training in the ALMA setting}\label{algo}
  \begin{algorithmic}[1]
    \Procedure{Train}{$m,w$, replay, grow}\Comment{$m$ is the model, $w$ is the waiting time}
    \State $t \leftarrow 1$
    \State $\mathcal{D} \leftarrow \emptyset$
    \While{$t<B$} \Comment{For each stage}
      \If{replay} \Comment{Acquire $w$ mega-batches}
        \State $\mathcal{D}\leftarrow\mathcal{D} \cup \mathcal{D}_t \cup ... \cup  \mathcal{D}_{t+w-1}$
      \Else
        \State $\mathcal{D}\leftarrow  \mathcal{D}_t \cup ... \cup  \mathcal{D}_{t+w-1}$
      \EndIf
      \State $t \leftarrow t+w$
      \If{grow}
        \State $m.grow()$ \Comment{Grow the model if the model is a growing model}
      \EndIf
      \State $m.train(\mathcal{D})$ \Comment{Fine-tune or retrain from scratch $m$ on the collected dataset}
    \EndWhile
    \EndProcedure
  \end{algorithmic}
\end{algorithm}

\subsection{Fixed Architectures}
\label{sec:fixed}
The first family of methods trains models with a fixed architecture. These models are sequentially trained over new mega-batches and exhibit a fixed memory footprint. We consider three models:

\paragraph{Single Model (\Single): } This is a standard multi-layer neural network (e.g., fully connected neural network or transformer) trained by stochastic gradient descent and initialized from the parameters of the model trained on the previous mega-batch, unless otherwise specified.

\paragraph{Ensemble of Models (\Ens): } The second approach is the simplest modular approach, consisting of an ensemble of $N$ neural networks with the same architecture but different random initialization seed, each being trained independently on the same data.  
The output of the overall model at test time is the average probability distribution produced by each component\footnote{Classical bagging approaches and majority vote strategies have been also explored without significant difference.}. The advantage of \Ens{} is that training and inference can be trivially parallelized, enabling to scale up model parameters very easily. The disadvantange is that  inference requires $N$ times more compute than what is required by each component.

\paragraph{Uniform Mixture of Models (\UMix): } A potential drawback of \Ens{} is that evaluation and training are inconsistent, \textcolor{black}{meaning that training and testing use different model predictors}. \UMix{} addresses this by training a model whose prediction is the average (in logit space) of the predictions produced by $N$ networks. While this requires synchronization during training, now both training and evaluation use the same model.

\subsection{Growing Architectures}
\label{sec:grow}
In the previous section, the number of parameters and the architecture of the model are fixed throughout the model's lifetime. However, as more data is observed, it is interesting to consider dynamic architectures that grow over time, because these may save compute and memory during the earlier stages of learning while providing more predictive power during the later stages. We consider three  growing approaches:

\paragraph{Growing Ensemble (\gEns): } Like the \Ens{} model, \gEns{} is also a combination of neural networks trained independently. While  \Ens{} considers a fixed number of networks that are, at each stage, trained over the new chunck of data, \gEns{} replaces this step by a growing step where $k$ new neural networks are added. In our implementation, only these $k$ neural networks are trained over the new data, while the other neural networks (trained on previous mega-batches) are kept fixed. Therefore, when starting with a single component and until the next growing step, the cost of training \gEns\ is equal to \Single\ for the same model architecture. Unless otherwise specified, we use $k=1$ for the experiments in the paper.

\paragraph{Growing Mixture of Experts (\gMoE):} A hierarchical mixture of experts models (MoE) is an architecture where at layer $l$ the output representation $z^l$ is: 
$z^l = \sum_{j=1}^k g(j | z^{l-1}) h(z^{l-1} | j)$, where $g$ is the gating or routing function and $h(\cdot | j)$ is the $j$-th expert. Compared to \Ens, MoE has exponentially many more components albeit with a lot of parameter sharing. Another advantage is that when selecting only one (or a few) experts, the computational cost is independent of the number of experts, assuming the cost of gating is negligible compared to the cost or executing the experts. The main issue is that MoE are notoriously harder to train~\citep{eigen14, denoyer15, gshard}. In this work, we consider a growing version of MoE, which we denote with \gMoE, whereby experts are added gradually over  time.
This has a tree structured gating function where leaves correspond to experts. At each layer, we calculate each expert's contribution to the total loss by summing the losses of the examples routed through that expert. We then "split" the expert responsible for the largest contribution to the loss. The split is performed by adding an expert with the same parameters, and turning the corresponding leaf node of the gate into a binary internal node with a child leaf for the old and new expert. This process guarantees that right before and right after a growth step the loss is the same. See Appendix~\ref{agmoe} for further details.

\paragraph{Firefly~\citep{firefly} (\FF):} \FF{} is a method which progressively grows neural networks, jointly optimizing both the model architecture and parameters. Growth includes both a width expansion by adding new hidden units (or feature maps) as well as a depth expansion by adding new layers. Importantly, this is an example of non-modular method unlike \Ens{} or \gMoE, which is potentially more expressive but also more inefficient at inference time because there is no structured sparsity that can be leveraged to speed up computation.

\section{Experiments} \label{sec:experiments}
\vspace{-7pt}

In this section we first describe how standard benchmarks can be repurposed for ALMA, we then provide the details of the models we tested, and we finally conclude
with an analysis of the results we obtained, aiming to understand which method attains the best trade-off between time, accuracy, compute and memory usage.

\paragraph{Datasets}
We consider a variety of datasets.
The first dataset is CIFAR~10~\citep{cifar10} that has a training set with
$50{,}000$ images of size $32$x$32$ pixels belonging to $10$ classes such as
bird, car, horse, ship, truck, etc. 
The second dataset is MNIST~\citep{mnist},
which consists of a training set with $60{,}000$ quasi-binary handwritten digits of
size $28$x$28$ pixels, and a test set with $10{,}000$ examples.
The third dataset, used for our large-scale language modeling evaluation, is a portion of the collection of English language text introduced in \citet{liu2019roberta}, consisting of Books, Wikipedia and Common Crawl. 
We consider 4 (large) mega-batches for training and one additional mega-batch for evaluation, each consisting of approximately 440M words; we also hold out a validation set with approximately 0.5M words of Common Crawl for model selection.
We use a byte-pair encoding (BPE)~\citep{sennrich2016neural} vocabulary with $50{,}000$ units, following~\citet{radford2019language}. This dataset is fairly representative of what practitioners might face when maintaining a deployed system with new data arriving every few months.


Given a dataset like any of the above, we construct a benchmark for
ALMA evaluation as follows: 1) we randomly partition the training set
into $B$ mega-batches with equal number of training examples ($B=100$ for CIFAR, $B=500$ for MNIST and $B=4$ for the text dataset), 2) from
each mega-batch we extract $10$\% of the data to build the mega-batch
validation set (except for the large scale language modeling dataset where we use the provided validation set), and 3) we create a learning experience by doing one
pass over the sequence of mega-batches. For each mega-batch, the learner
can query as many mini-batches as desired. The learner can also decide
not to train on the data of a mega-batch right away but
instead {\em to wait} and accumulate data across a few consecutive
mega-batches. While the learner observes data, it is also tested on
the test set. This is not used for validation purposes, but only for
final reporting as shown in \textsection\ref{sec:results}. 

\paragraph{Models}
We evaluate the six approaches presented in \textsection\ref{sec:algo}, and for each of them we consider various waiting times, a version with and without replay, and at least four model sizes. 
For methods with expanding architectures, we try different configurations of hyper-parameters controlling \textit{when} to grow, and \textit{how much} to grow. For simplicity, we limit expansion phases to occur in between megabatches. Next, we describe in details the architecture used on each dataset. Further experimental details to aide reproducibility are reported in Appendix~\ref{app:hyperparams}.
On MNIST 
the backbone architecture of \Single{} is a three layer fully connected neural network with
ReLU units. We considered various hidden units size, ranging from 4 to 64 (which we refer to as  [small] and [large], respectively), which let us simulate the regime of big data relative to the size of the network and explore how to grow architectures without worrying about overfitting.  
Similarly, the components of \Ens, \gEns{} and \UMix{} are \Single{} networks of the same size as stated above; \gMoE{} also starts off as \Single{} and adds modules (at the first two layers) that have the same size as the original layer of \Single.  

On CIFAR~10, the methods and notations are the same as in MNIST. The only difference is that the backbone architecture is a 
scaled down version of a VGG19 convolutional neural network~\citep{vgg} as in \citep{firefly}, where the
number of intermediate feature maps is the same for each layer, ranging from $4$ to $64$. On this dataset, we also consider \FF{} starting off from the same VGG19 backbone.

For the language modeling task \Single \ is a Switch Transformer~\citep{fedus2021switch},
which is a hard mixture of experts model with an additional load balancing loss term and hard capacity constraint applied during training to prevent uneven expert utilization.
Following~\citet{fedus2021switch}, we fix the weight of the balancing loss term to $0.01$ and use a capacity factor of $1$, ensuring relatively uniform expert utilization.
We train the model using Adam~\citep{DBLP:journals/corr/KingmaB14} and tune the learning rate and dropout on the validation set.
In the growing setting we copy the expert weights and gating network weights corresponding to the top-$k$ experts incurring the largest loss, where $k$ is typically between 2 and 4. This growing procedure preserves a flat mixture and adds multiple experts at the time. While this approach performs slightly worse than the one described in \textsection\ref{sec:grow}, it is easier to implement at scale.
We consider two model sizes: a \emph{base} model with 6 layers and model dimension of 512, for a total of 40M shared parameters and 6M additional parameters per expert; and a \emph{large} model with 12 layers and model dimension of 768, for a total of 96M shared parameters and 28M additional parameters per expert. We use an input sequence length of 512 tokens and we do not use replay given the large \textcolor{black}{mega-batch} sizes. During each mega-batch, we train all language models for exactly 120000 gradient steps (results in Fig. \ref{fig:LM}) unless otherwise specified (e.g. Tab. \ref{tab:seq_abl}). This makes it easier to compare models for the same computational budget at the \textcolor{black}{mega-batch} level.

\section{Results} \label{sec:results}



\subsection{Visual Recognition}
\begin{figure*}[t]
  \centering
\includegraphics[width=.98\textwidth]{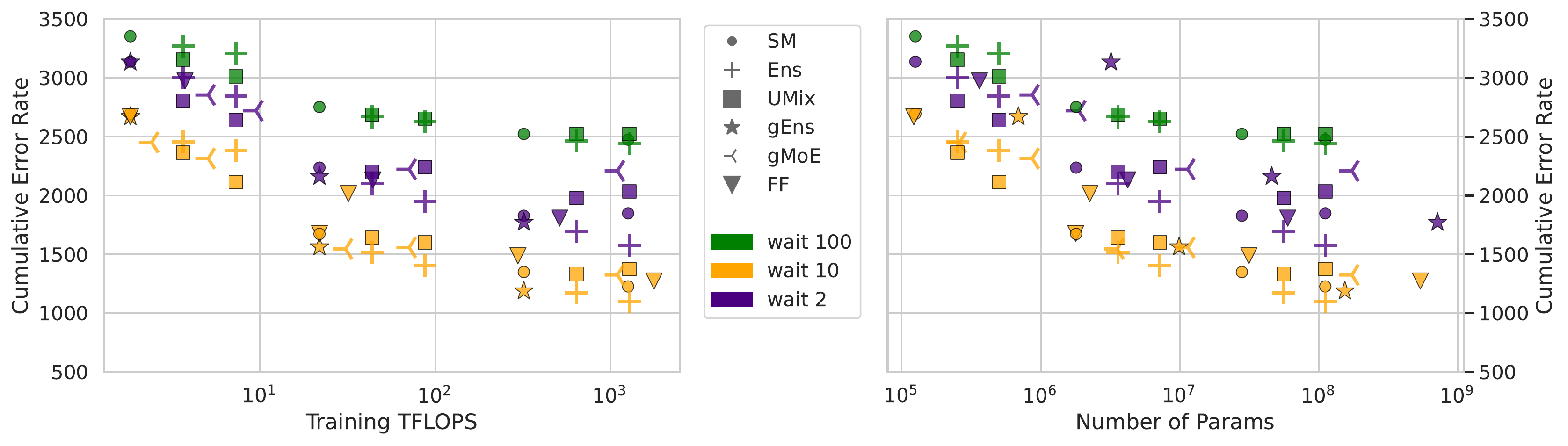}
  \vspace{-10pt}
  \caption{\small CIFAR~10 results: Cumulative error rate versus cumulative flops and number of parameters without replay.
    For the same model type we vary the size of the backbone architecture and the waiting time.}
  \label{fig:cifar_CER}
\end{figure*}

Since conclusions are rather similar, we focus our analysis on the more challenging CIFAR~10 dataset, and report results also on MNIST in Appendix~\ref{app:vision}.

\begin{figure}[t]
 \vspace{-5pt}
  \centering
\includegraphics[width=.98\textwidth]{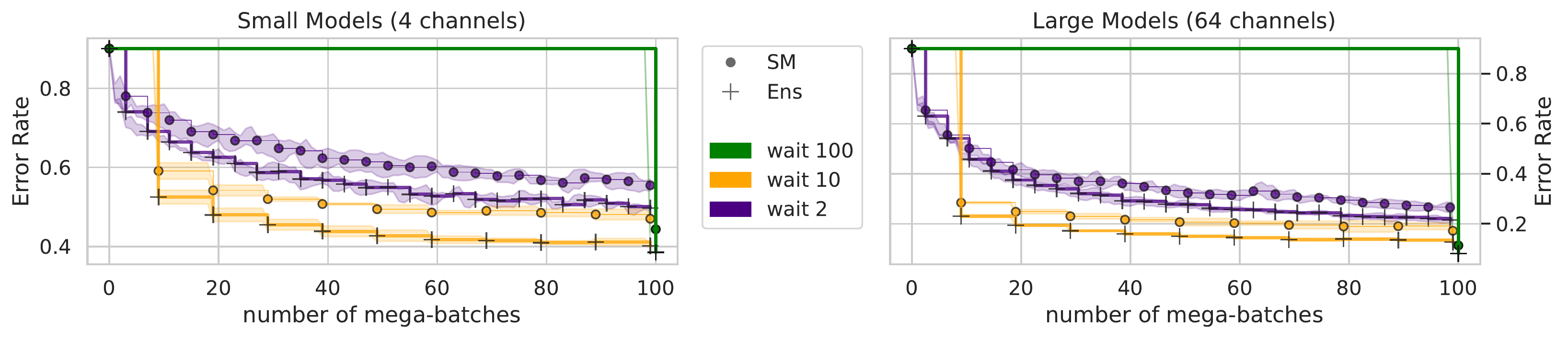}
\includegraphics[width=.98\textwidth]{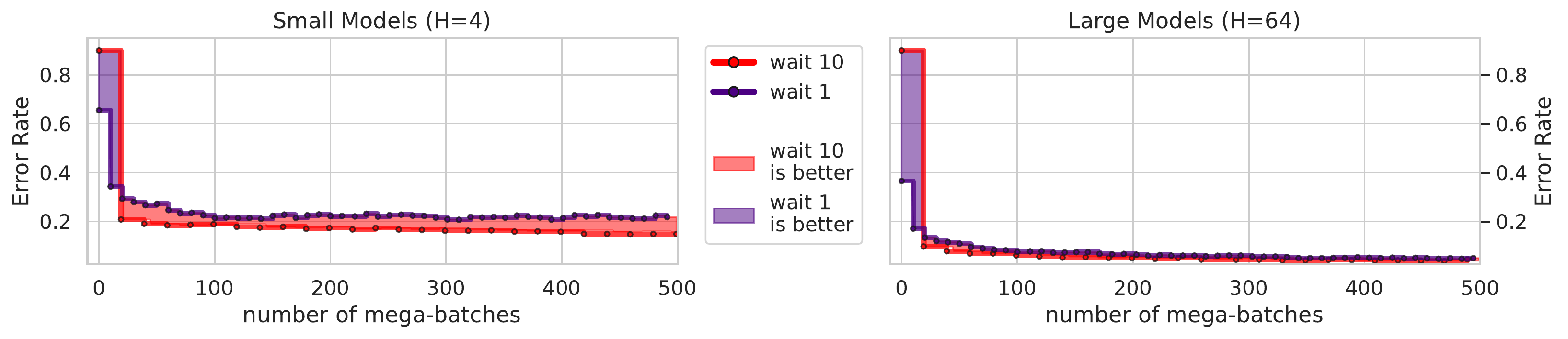}
  \vspace{-10pt}
  \caption{\small Error rate over time of small models (left) and large models (right) on CIFAR~10 (top) and MNIST (bottom).}
  \label{fig:cifarmnsit_anytime}
  \vspace{-3pt}
\end{figure}

\textbf{Smallest waiting time might not be optimal:} \textcolor{black}{We begin our analysis in the setting without replay, shown in } Fig.~\ref{fig:cifar_CER}. We first observe that an intermediate waiting time (in this case equal to 10) strikes the best trade-off between Cumulative Error Rate (CER) and both training cost (left) and memory cost (right). As shown in Fig.~\ref{fig:cifarmnsit_anytime}-top, where the test error rate is plotted as a function of the number of mega-batches received, greedy methods using waiting time equal to 2 achieve a lower error rate only during the very beginning of the stream, but are outperformed later on. Tardy predictors waiting for all 100 mega-batches before training obtain the best final accuracy, but have no predictive capabilities throughout the first 99 mega-batches. Instead, methods with an intermediate waiting time (shown in orange) can quickly deliver a reasonable predictor early in the stream, and obtain a final error rate that is very close to the lower bound obtained by tardy methods. Thus, a waiting time of 10 yields the lowest area under the curve (or CER) on CIFAR~10. 

On MNIST however, an intermediate waiting time is best only for small models, as shown in Fig. \ref{fig:cifarmnsit_anytime}-bottom. Very greedy models do not converge as well in this setting, which leads to a significant penalty in terms of CER. However, bigger networks converge very fast in just a few megabatches, making smaller waiting times more desirable. Therefore, the optimal waiting time depends on several factors such as the model size, the time horizon, how difficult the task is and how quickly the model learns. In such non-convex setting, it is certainly not necessarily true that learning on the data as soon as it becomes available attains always the best trade-off between error rate and compute.

\textbf{Larger models are more statistically efficient:} Second, we observe that bigger models (\Single{} and \Ens{}) not only generalize better but they are also statistically more efficient: on the small \Ens{} obtained almost 40\% error rate by the end of its learning experience (Fig.~\ref{fig:cifarmnsit_anytime}-top left), which is worse than the error rate obtained by the large \Ens{} just after having observed one tenth of the entire stream. The statistical efficiency of large models does not apply only to large transformers~\citep{DBLP:journals/corr/abs-2001-08361}, but also to fully connected (we obtained similar results on MNIST, see Fig.~\ref{fig:cifarmnsit_anytime}-bottom) and convolutional models.
 
\textbf{Growing does not improve: } If we focus our attention on the three approaches with growing architectures, namely \gMoE{}, \gEns{}, and \FF, we find that there is no clear winner among them. When comparing across a fixed computational budget (Fig.~\ref{fig:cifar_CER} left), \gEns{} overall performs better than \gMoE{} and \FF{}. However, when we fix the memory budget instead (Fig.~\ref{fig:cifar_CER} right), \gEns{} is, on average the worst performing method.

Next, we investigate the efficiency of growing, since in principle, we would expect that adapting model capacity with the amount of data should strike a better trade-off between accuracy and memory/compute usage. For a fixed computation or memory budget, it is always better to \textbf{start with a larger model}, rather than growing it over time. Indeed, we find that on both graphs of Fig.~\ref{fig:cifar_CER}, \Single{} almost always outperforms \gMoE{} and \FF{}, a trend that is especially visible for higher budgets of TFLOPS and parameters. In other words, a \gMoE{} or \FF{} that starts small and finishes big will typically be outperformed by a \Single{} model of average size.

Finally, \Ens{} is more efficient than \gEns{} in terms of memory, but vice versa in terms of training compute. However, should we look at the inference cost of both methods, we would find that \Ens{} outperforms its growing counterpart, whose inference cost grows over time while it is fixed for \Ens{}. Once again, the best strategy is to pick the largest fixed-capacity model for a given computational budget. Notice that these conclusions apply to the methods considered in this study, and improving approaches that dynamically adapt their architecture over time is clearly a worthwhile avenue of future research. 
 
\textbf{Operating point matters: } 
We proceed by contrasting \UMix{} and \Ens, where the former averages predictions during training between different components, while the latter trains each component independently. In all our experiments when working with smaller models, \UMix{} has a slight edge on both memory and compute fronts; however as the size of each component gets bigger the trend reverses, and \Ens{} outperfoms \UMix.
We surmise that smaller models suffer the most from the inherent inefficiency of ensembling which forces each component to learn the same set of features. When capacity is limited, it is better to coordinate learning among the components instead. 
Overall, this finding highlights how conclusions about which model works best really depends on the operating point. Only when we consider the full spectrum of model sizes, can we conclude which approach works best.

\textbf{\Ens{} strikes the best trade-off: } More generally, {\bf \Ens{} is the best performing method for larger models} across all our experiments, including the language models reported in \textsection\ref{sec:lm}. This is a remarkable finding given the simplicty of the approach and how easy it is to parallelize their training process. Ensembling makes it very easy to increase model capacity early on, and it is so far the best way to utilize compute at scale, a possible indication of the inefficiency of training large models using alternative approaches, which highlights yet another worthwhile avenue of future research.


\begin{wrapfigure}[17]{R}{0.5\textwidth}
\centering
\vspace{-10pt}
\includegraphics[width=0.48\textwidth]{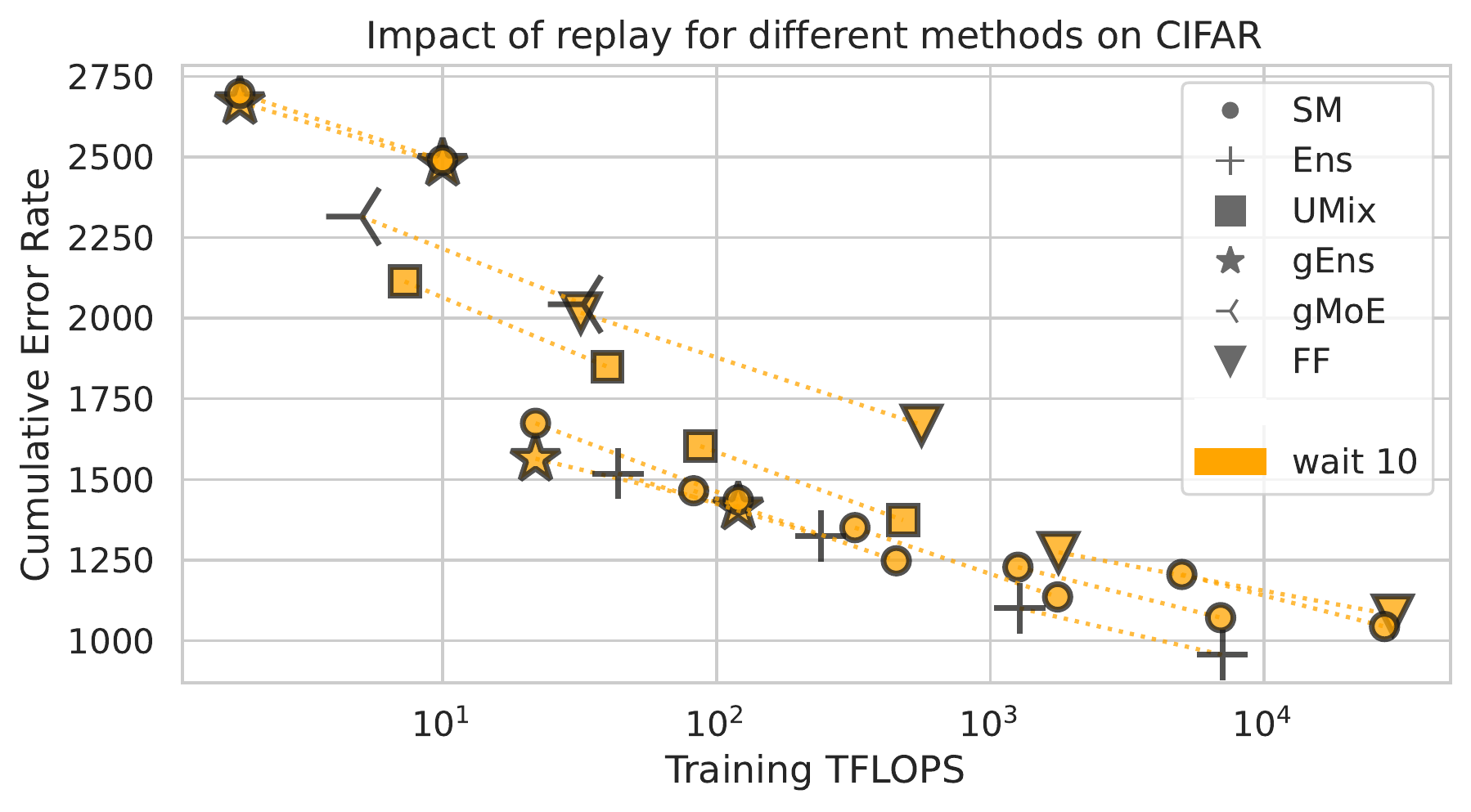}
\caption{Impact of replay on the CIFAR-10 dataset with a wait time of 10.
  For each method we show a line from the result without replay (left) and with replay (right).
  }
\label{fig:replay_cifar}
\vspace{15pt}
\end{wrapfigure}

\textbf{Replaying past mega-batches does not improve:} We now consider the same approaches as before but with models trained on all megabatches seen so far. Therefore, at the very last training step, models are trained on the entire dataset (concatenation of all megabatches). In Fig. \ref{fig:replay_cifar} we report the results when the waiting time is equal to $10$.
In all cases, replaying data gives better results at the expense of an increase in compute. Except for \gEns{}, these gains are roughly the same for all methods, as all segments are parallel to each other. \gEns{} gains less as the last component which is trained on the full dataset has disproportionate influence in the model average which includes components trained on fewer megabatches. However, this last component essentially coincides with \Single{} trained on the full dataset. Hence the two methods converge to the same performance when using replay. We provide additional results with replay in Appendix \ref{sec:cifar_replay}, \textcolor{black}{which shows that there are benefits from replaying only at higher computational budgets where also the optimal waiting time reduces to $1$}. 

More importantly, we observe that {\bf replay does not yield a significantly better trade-off} between CER and compute. For the same computational budget, methods using replay attain similar CER of methods that do not use replay. Other factors such as the size of the backbone architecture or the waiting time matter more.

\subsection{Language Modeling Experiments} \label{sec:lm}

For the large-scale language modeling experiments, we consider two model sizes (base and large, see \textsection\ref{sec:experiments}),  with an inference cost per input of 42 and 126 GFLOPS, respectively.
The number of experts is set to 4, 8 and 12 for \Single{}, and it does not affect the inference cost since only one expert per input is selected regardless of the total number of experts.
Due to the computational burden of these experiments (in total more than 200 GPU days), we limit our analysis to four mega-batches.
Nevertheless, this scale (of model and data) and type of application are rather representative of a typical ALMA setting.  
Please refer to Tab.~\ref{tab:lm} in Appendix~\ref{app:lm} for a comprehensive analysis, as here we will only highlight the major findings.

\begin{figure*}[t]
  \centering
\includegraphics[width=1\textwidth]{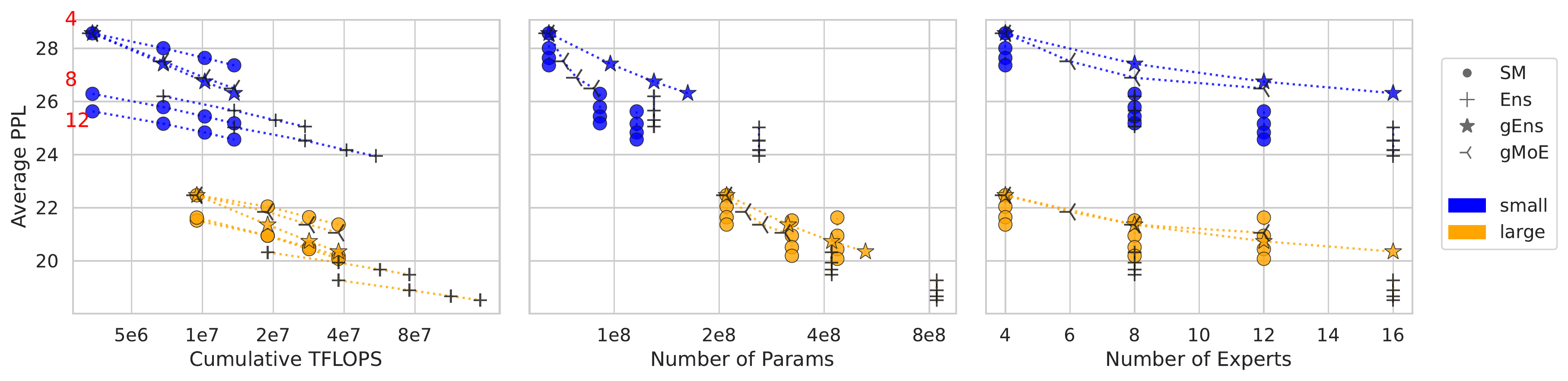}
  \caption{\small Language modeling trade-offs: average perplexity (PPL) versus cumulative compute, number of parameters and number of experts. Numbers in \textcolor{black}{red} refer to the number of experts in the corresponding \Single{} runs.}
  \label{fig:LM}
\end{figure*}

The main results are presented in Fig.~\ref{fig:LM}. Each line is a trajectory with four points, one for each mega-batch in the stream, as we report average as opposed to cumulative perplexity.   
For a given model size and for a given computational budget, there are three \Single{} models, one for each number of experts we consider, namely 4, 8 and 12.

\textbf{Larger models are more efficient: } In agreement with our results on computer vision tasks, we observe that bigger models tend to generalize better and are more sample efficient. For instance, the large model after a single mega-batch outperforms all base models, including base models after four mega-batches which have seen four times more data. This finding is consistent across all methods tried for this experiment. 

\textbf{Growing does not improve: } Once again, there is no clear winner among growing methods. For larger models, \gEns{} outperforms \gMoE{} for the same compute, and perform similarly for base models. However, for all model sizes, \gMoE{} is more memory efficient, therefore the optimal approach among them will depend on both compute and memory budget. More importantly, we observe that models with fixed capacity are more compute and memory efficient than models that grow over time. Looking at the average perplexity as a function of the number of experts, we see that methods which start with a small number of experts and later grow are outperformed by similar fixed architecture which have an intermediate number of experts. This highlights the importance of having more capacity at the start of training, rather than at the end.

\textbf{Ensembles perform the best: } Third, \Ens{} thrives in the larger capacity setting. Looking at the orange markers in the graph, we see that for equal computation budget, \Ens{} methods outperform all other methods, which is consistent with the computer vision results. In the base setting instead, versions of \Single{} (see the lowest blue points) strike a better tradeoff in both compute and memory. 



\textbf{Learning sequentially is harder: } We argued initially that once the learner makes several passes over each megabatch, the data distribution cannot be considered i.i.d. anymore, relative to the empirical distribution of the union of all megabatches. It is however unclear how much this issue has a practical impact in the performance of the model. 
In order to assess this we run one last experiment using our best performing approach, namely \Ens. We compare a model trained on $k$ mega-batches sequentially with the same model trained all at once on the aggregation of the same $k$ mega-batches. Since both approaches have the same computation budget, the same architecture and are fed with the same data, we can disentangle the effect of the non-i.i.d nature of the data in ALMA.
The results shown in Tab.~\ref{tab:seq_abl} confirm that ALMA's sequential (seq.) training is indeed more challenging. Across all four configurations, models incur a drop in performance when compared to regular i.i.d training, and even more so when the model is larger.
This gap offers another opportunity of future research on ways to make sequential training more effective when using deep non-linear neural networks.

\begin{table}[t]
\small
\centering
\begin{tabular}{ccccccc}

\toprule
Method  &  PPL $k=3$, iid & PPL $k=3$, seq.  &  & & PPL $k=4$, iid & PPL $k=4$, seq.  \\
\midrule
Small \Ens{} 4@2   & \textbf{24.30} & 24.57 &  & & \textbf{24.13} & 24.35 \\
Big \Ens{} 4@2     & \textbf{18.04} & 19.14 &  & & \textbf{17.88} & 18.92 \\
\bottomrule

\end{tabular}
\vspace{4pt}
\caption{Ablation on the effect of learning sequentially (seq.) as opposed to learning with fully i.i.d. data, for the same amount of data and compute. The model is an ensemble with 2 components each of which with 4 experts per block.}
\label{tab:seq_abl}
\end{table}





%
\vspace{-.1cm}
\section{Conclusions} \label{sec:remarks}
In the abstract we promised the reader to provide an empircal answer to several questions:

1) {\em How long should the learner wait before training on the newly arrived \textcolor{black}{mega-batches}?} There is no single answer to this question. We have seen that on CIFAR~10 but also on MNIST when using smaller architectures \textcolor{black}{and when using replay with smaller compute budgets}, an intermediate waiting time strike the best trade-off. However, there is no known formula for deriving the waiting time, as it depends on several factors such as the time horizon, the initial performance of the model and how quickly a model learns, to name a few. The firm conclusion is that greedily updating the model as soon as data becomes available, as advocated by literature on convex online learning, might not always be the best strategy when using deep neural networks, In practice, also waiting too long, to the point that the learner does not even have time to perform a single pass over the aggregated mega-batches, might be suboptimal.

2) {\em What architecture should the learner adopt?}
Our study indicates that, among all methods we tested, ensembling strikes the best trade-off in general. Ensembling is simple and easily parallelizable, and it offers a straightforward way to increase capacity. Starting off with a larger model, for instance via ensembling, is an excellent way to obtain good anytime performance.

3) {\em Should the learner increase capacity over time as more data is observed?}
The answer is negative, currently. It is better to start off with the largest architecture fitting into memory and keeping that fixed.
A cynical interpretation of this conclusion could make the reader believe that growing the architecture size should not be a topic of interest. However, as data is added over time so is computation and memory. It is often the case that researchers working on large-scale 
learning instantiate (rightly so) the biggest possible model to train on their task, but few months later they can manage to launch even bigger models thanks to compute and engineering advances. How can the larger model leverage what has been learned from the previously trained model? Is there a modeling choice that strikes a better trade-off than retraining from scratch? More generally, what are good approaches to extract information from a new batch of data to integrate it into an existing model? We believe these are great avenues of future research, and that our ALMA framework (learning and evaluation protocol, codebase, baselines) provides a good abstraction of the practical setting, and a sound tool to pursue such investigation.

\section{Reproducibility Statement}
We have made several efforts to ensure that the results provided in the paper are fully reproducible.
We first provide an easy-to-use codebase from which all the computer vision results in this paper are generated. In this codebase, one can find the exact hyperparameters used for each method in the provided configurations. We have attached a readme to the code in order to guide users running our code. For the LM experiments, as stated in the appendix we use the fairseq~\citep{ott2019fairseq} and provide the required information to replicate our results. 

\section{Acknowledgements}
 We would like to thank Csaba Szepesvari for discussing how ALMA  relates  to online learning, J{\"o}rg Bornschein for general discussion and for pointing out at missing experimental results, and Thang Doan for giving feedback on earlier drafts.




\bibliography{collas2022_conference}
\bibliographystyle{collas2022_conference}
\newpage
\appendix
\section*{Appendix}

\section{Growing Mixtures of Experts}
\label{agmoe}
\begin{wrapfigure}{r}{0.5\textwidth}
  \begin{center}
    \includegraphics[width=0.5\textwidth]{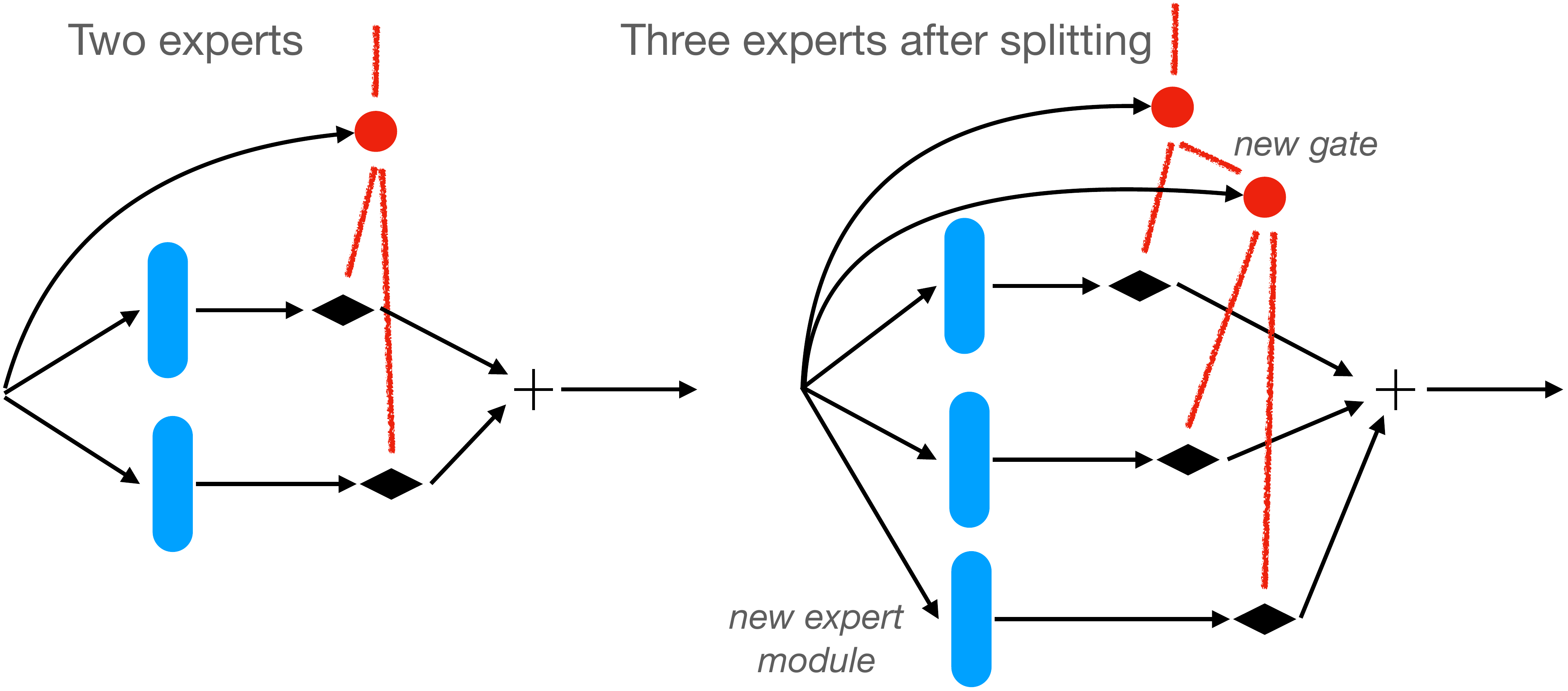}
  \end{center}
  \vspace{-.5cm}
    \caption{\small Illustration of a growth step in a tree structured
    mixture of experts. A network is composed of several layers
    like this. The blue squares are experts (e.g VGG layers). The red
    elements corresponds to the gatings which, given an input compute
    a score for each expert. When splitting an expert (right), the
    gating structure is updated by creating a child gate, and an
    additional expert is added to the mixture.}
\label{fig:toy_gmoe}
\end{wrapfigure}

\paragraph{Growing Mixture of Experts (\gMoE):} A mixture of expert (\MoE) is a sequence 
of non-linear functions, each of which is potentially a mixture of
experts (omitting the dependence on parameters):
\begin{equation*}
  m(x) = f^{l}(f^{l-1}(\dots f^1(x) \dots)), \text{ with }  f^i(z)  =  \sum_{j=1}^k g^i(j | z) h^i(z | j)
\end{equation*}
where $g^i$ is the gating function at the $i$-th layer which outputs a
categorical distribution over the number of experts, and $h^i(\cdot
| j)$ is the $j$ expert at layer $i$. The gating function can be ``soft'' in which case it outputs non-zero weights for each expert via a softmax, or ``hard'' version in which case only one expert is selected through a multinomial sampling (and learned through the straight-through estimator in this paper ~\citep{DBLP:journals/corr/BengioLC13}). At test time in the ``hard'' case, we select the expert
with the largest probability. The interest of mixtures of experts is they have a high expressivity, and experts can be easily added to increase the capacity of the model. The \gMoE{}  model is the growing version where, at each stage as illustrated in Fig. \ref{fig:toy_gmoe}, new experts are added at each layer~\citep{gmm_exp2}. 

The key design considerations are: {\em when} to grow,
{\em what} to grow and {\em how} to grow. 
Here, we
will refer to our default setting which favors simplicity, unless otherwise specified.

A growth step is triggered at each stage, ensuring a linear growth over time.
We grow by adding one expert at each layer, making sure that all experts within a layer have
the same architecture albeit with different parameters. In order to grow, we look at which expert has associated the largest cumulative loss; we call such expert the {\em
  losing} expert. The cumulative loss is defined as the sum of
the losses of examples on the validation set that have been routed
through a particular expert; each expert has associated a cumulative
loss value. The rationale is to identify at each layer the expert responsible for the largest contribution to the total loss.

To avoid drop in the loss function and to keep its differentiability when splitting an expert, we propose a tree-based approach where the losing expert is split into two 
experts with exactly the same parameters as illustrated in Fig.~\ref{fig:toy_gmoe}: Two children leaves are derived and we instantiate a new gating
for the children which decides  whether an input
example routed to the old expert, should now go to the right or left
expert child. The parameters of the new gate
are initialized at random while the parameters of the new experts are exact copies of the ones of the losing expert that we split.  

More formally, if $s$ is the losing expert
then the term $  g^i(s | z) h^i(z | s) $ is replaced by:
\begin{equation}
\sum_{k=1}^2 g^i(s | z) g^i(k | z, s)  h^i(z | s, k) \label{eq:hmoe_term}
\end{equation}
where $g^i(k | z, s)$ is the newly introduced gate, and $z$ is the
input of the gating and experts.

Over time, the gating function learns to
partition its input space into a binary tree (if we start from a
single expert), and the gating value of an expert is the product of
the gating probabilities on the path from root to the leaf expert.
Both the gating tree structure and the particular initialization scheme  guarantee that the
growth step is smooth and fully differentiable, in particular, the
loss before and after the growth step is the same.

If we consider each path in the MoE model to be a different model, then with $L$ layer of $k$ MoE components, there are $k^L$ many possible paths through the MoE model, hence the number of paths grows exponentially with the number of layers. You can think of this as an ensemble with exponentially many components, but this is still tractable because components share parameters.

 \begin{algorithm}[H]
    \caption{gMoE \label{algo:gMoE}}
    \begin{algorithmic}[1]
    \State $k$: number of mega-batches to aggregate
    \State  $\mathcal{D} = \emptyset$
    \Function{train}{$\mathcal{D}_i$, $i$}
        \State $\mathcal{D} \mathrel{+}= \mathcal{D}_i$
        \If{$i \mbox{ mod } k == 0$}
            \State Extract $\mathcal{D^{\mbox{VAL}}}$ and $\mathcal{D^{\mbox{TR}}}$ from $\mathcal{D}$
            \While{$m$ is not converged:}
               \State $(x,y) \sim \mathcal{D^{\mbox{TR}}}$
               \Comment In practice, sample   mini-batches.
               \State m.update($x,y$)
            \EndWhile
            \State $\mathcal{D} = \emptyset$
            \State m.grow($\mathcal{D^{\mbox{VAL}}}$)
            \Comment Growth step can be done at a different rate too.
        \EndIf
        \EndFunction
        
        \Function{grow}{$\mathcal{D^{\mbox{VAL}}}$}
        \For{each layer in the network}
        \State Let $i$ be the losing expert on $\mathcal{D^{\mbox{VAL}}}$, i.e. the expert incurring the largest cumulative loss.
        \State Turn corresponding gating output in an internal node and derive 2 gate children
        \State Initialize the new experts by copying the parameters from the old parent expert.
        \State Initialize the new gating between the two siblings at random.
        \EndFor
        \EndFunction
    \end{algorithmic}
\end{algorithm}


\section{Hyper-parameter settings} \label{app:hyperparams}
\subsection{Computer Vision Experiments}
For each megabatch received, we keep 10\% of the data to perform cross-validation. All experiments are run on a single 16GB Quadro GP100 GPU. We apply data normalization for each dataset considered. A training minibatch size of 128 is used. $\UMix{}$ and $\Ens{}$ models have $N=5$ in all experiments. for $\gEns{}$, we train one model $n=1$ at every mega-batch, so the total number of models depends on the amount of mega-batches. For Firefly we use a growth rate of $0.25$, meaning that at every growth phase, we add approximately a quarter of the initial number of parameters. 

\subsubsection{MNIST}
Models are trained for 100 epochs, and we report results with soft gating. We use the AdaDelta (\cite{zeiler2012adadelta}) optimizer with default learning rate of 1. We use a MLP with 2 hidden layers of varying width (e.g. 4,8 or 32 neurons). 

\subsubsection{CIFAR-10 }
Models are also trained for 100 epochs with a learning rate of 0.01. We use Stochastic Gradient Descent with momentum value of 0.9 and weight decay of $1 \times 10^{-4}$. During training, we apply random horizontal flips and select random image crops with padding of 4 pixels. For the architecture, we use the same reduced VGG with batch normalization as prescribed in \cite{firefly}. All layers are initialized with the same number of channels (e.g. 4, 8, or 32 channels). 
For the Firefly experiments, we keep all the Firefly-specific hyperparameters to the default values suggested in the author's public codebase. We make one exception to this, namely we adapt the growth ratio to result in linear (rather than exponential) growth.

\subsection{Language Modeling Experiments}
All the language models are trained using fairseq~\citep{ott2019fairseq} with a maximum of eight 32GB GPUs (NVIDIA V100), optimized with Adam~\citep{DBLP:journals/corr/KingmaB14} using $\beta_1 = 0.9$, $\beta_2 = 0.98$, $\epsilon =$1e-8. The learning rate is warmed up over the first several hundred updates (between 500 and 4000) and then linearly decayed to 0 over the remaining updates, with a peak value tuned between 2e-4 and 5e-3. Models are trained up to 120,000 updates with local batch size of 8 sequences per GPU, with gradient accumulation as needed to achieve a total batch size of 192 sequences; each sequence has 512 tokens. We fix the Switch Transformer balancing loss term to $0.01$ and use a capacity factor of $1$, following~\citet{fedus2021switch}.

\section{Detailed Computer Vision Results} \label{app:vision}

\subsection{MNIST}

We show equivalent figures to the ones presented for CIFAR (e.g. Fig \ref{fig:cifar_CER} and \ref{fig:cifarmnsit_anytime}). We note for a given waiting time, different models rank similarly as in the CIFAR results. The main difference with the other computer vision dataset is on the optimal waiting time. As we saw in Fig. \ref{fig:cifarmnsit_anytime}, on MNIST a predictor obtains good performance using very few mega-batches, making small waiting time competitive. Nevertheless, we do see that in terms of final error rate, a small waiting time underperforms, especially for small models. 

\begin{figure*}[h!]
  \centering
\includegraphics[width=.98\textwidth]{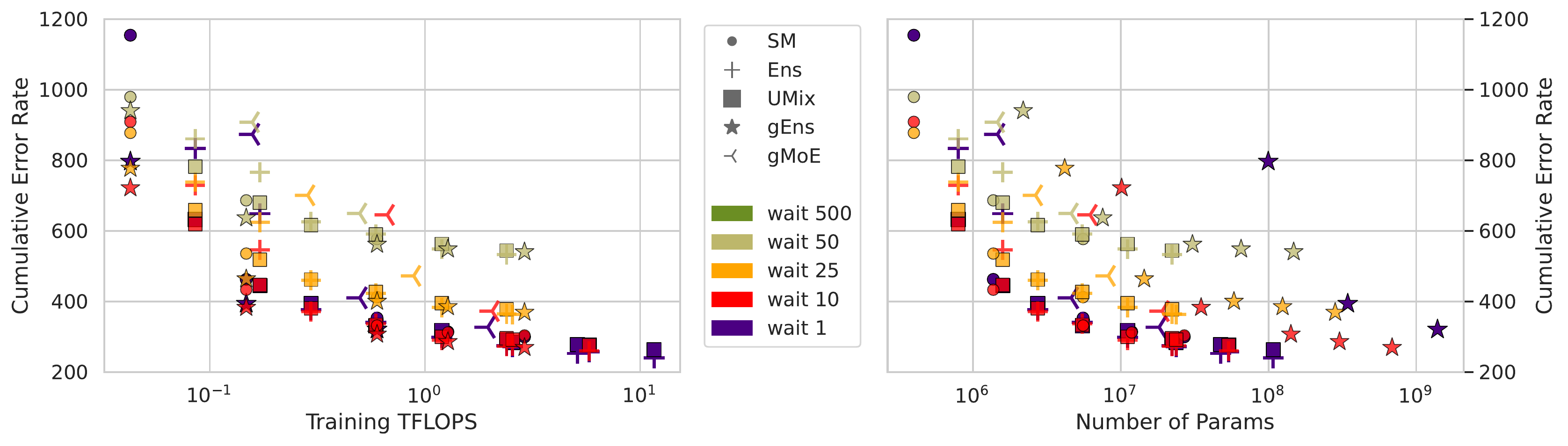}
\includegraphics[width=.98\textwidth]{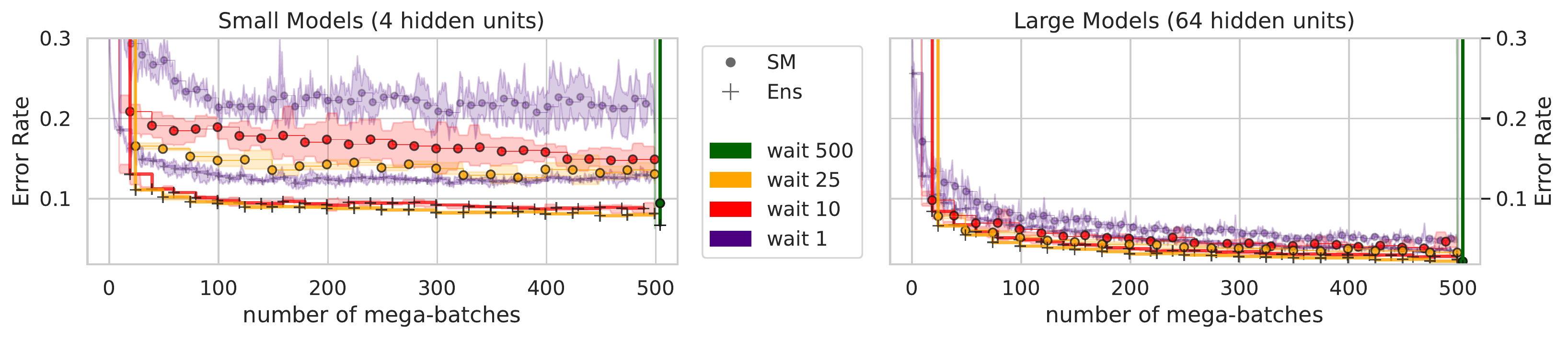}
  \vspace{-10pt}
  \caption{\small (top) Cumulative error rate versus cumulative flops and number of parameters without replay.
    For the same model type we vary the size of the backbone architecture and the waiting time. (bottom) Anytime Error Rate for the same methods on MNIST}
  \label{fig:mnist_both}
\end{figure*}

\newpage
\subsection{CIFAR-10}
\label{sec:cifar_replay}

For this dataset, we provide more results when using replay across a variety of methods and waiting times. We note in this setting, as the computational budget increases, the optimal waiting time decreases. This is because as more mega-batches is received, the training distribution gets closer to the ideal i.i.d scenario.
It can therefore bypass the optimization issues faced when training for multiple iterations on a small dataset. Again, we emphasize that this is not the case when using a small waiting time and no replay. 

\begin{figure*}[h!]
  \centering
\includegraphics[width=.45\textwidth]{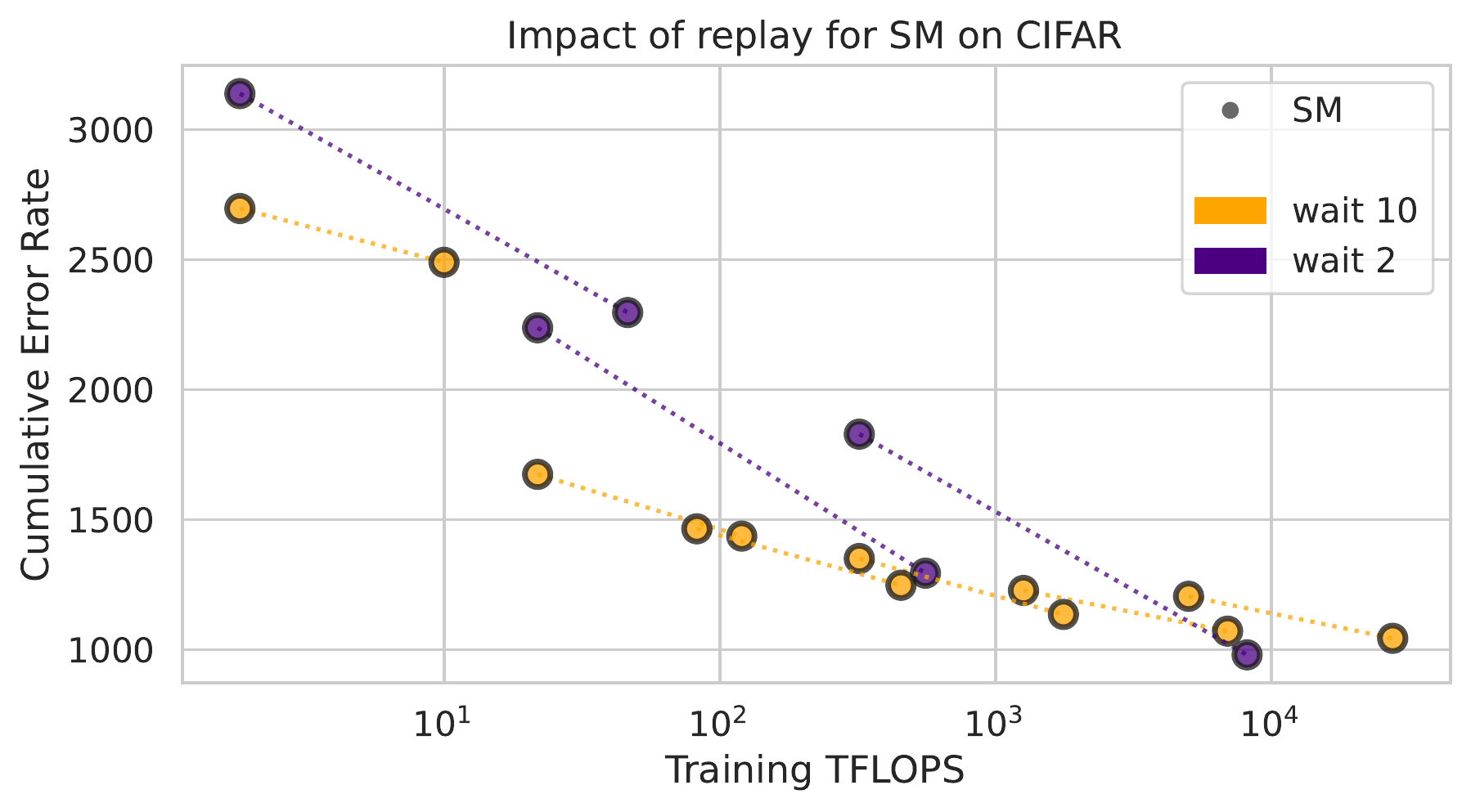} 
\includegraphics[width=.45\textwidth]{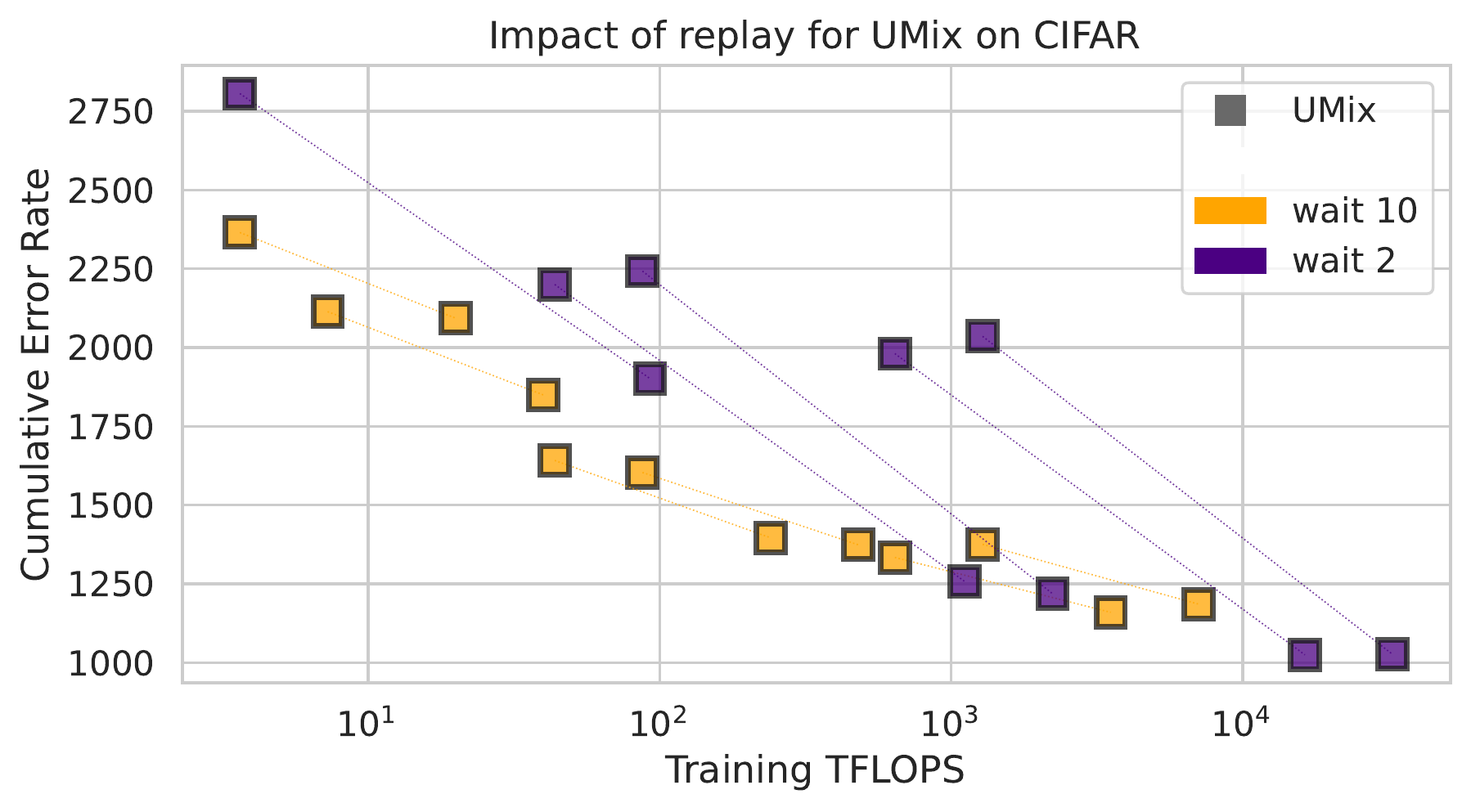}\\
\includegraphics[width=.45\textwidth]{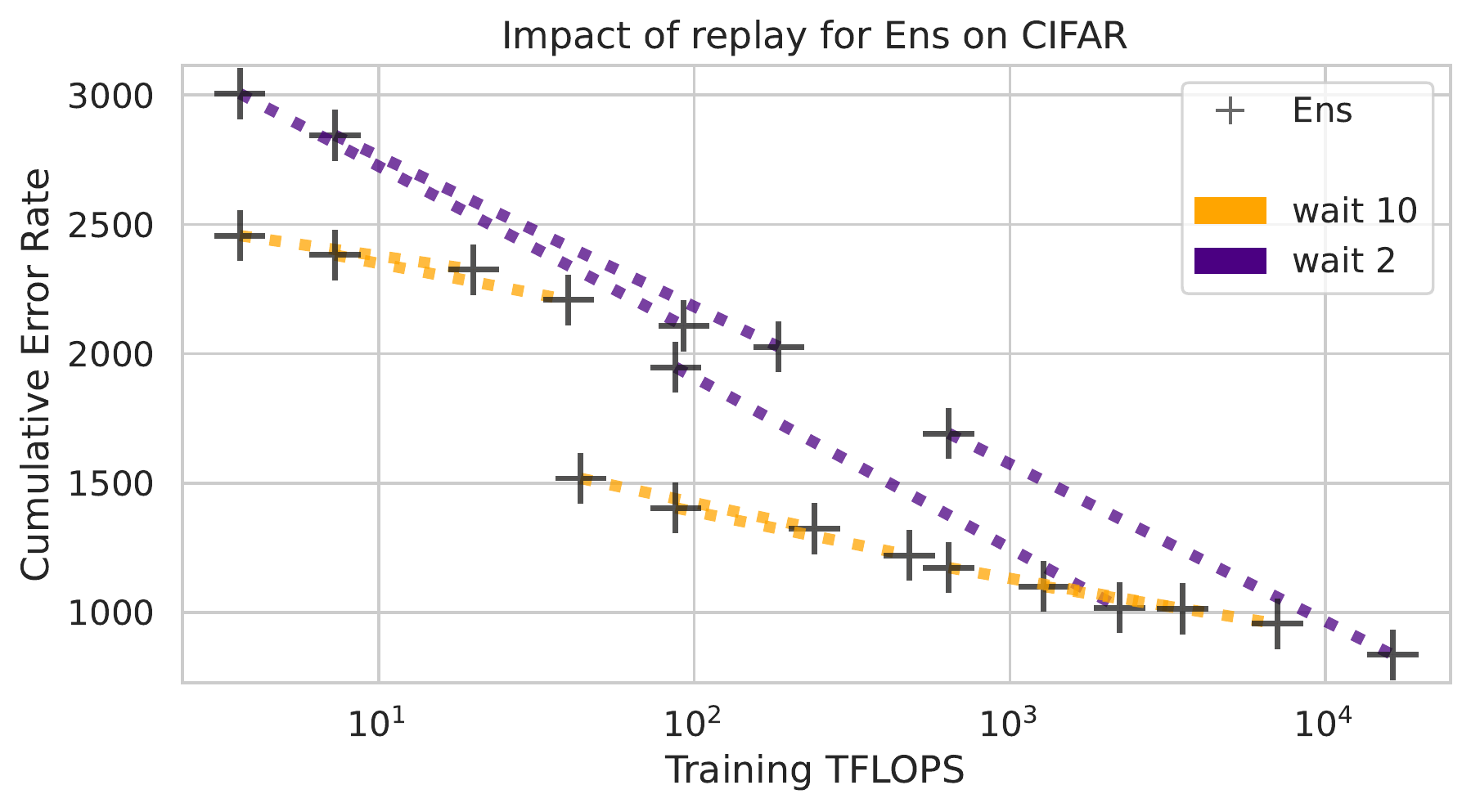}
\includegraphics[width=.45\textwidth]{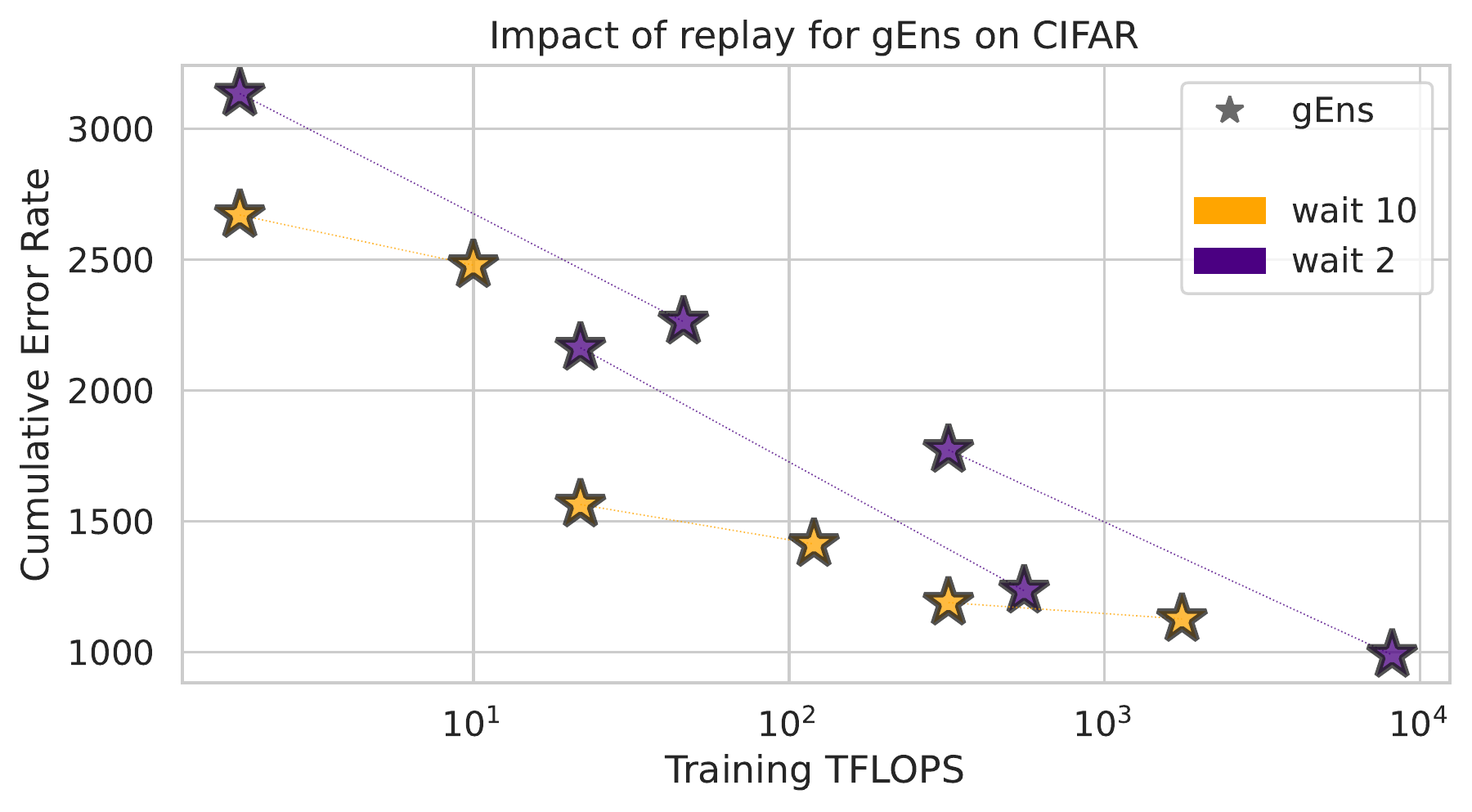} \\
\includegraphics[width=.45\textwidth]{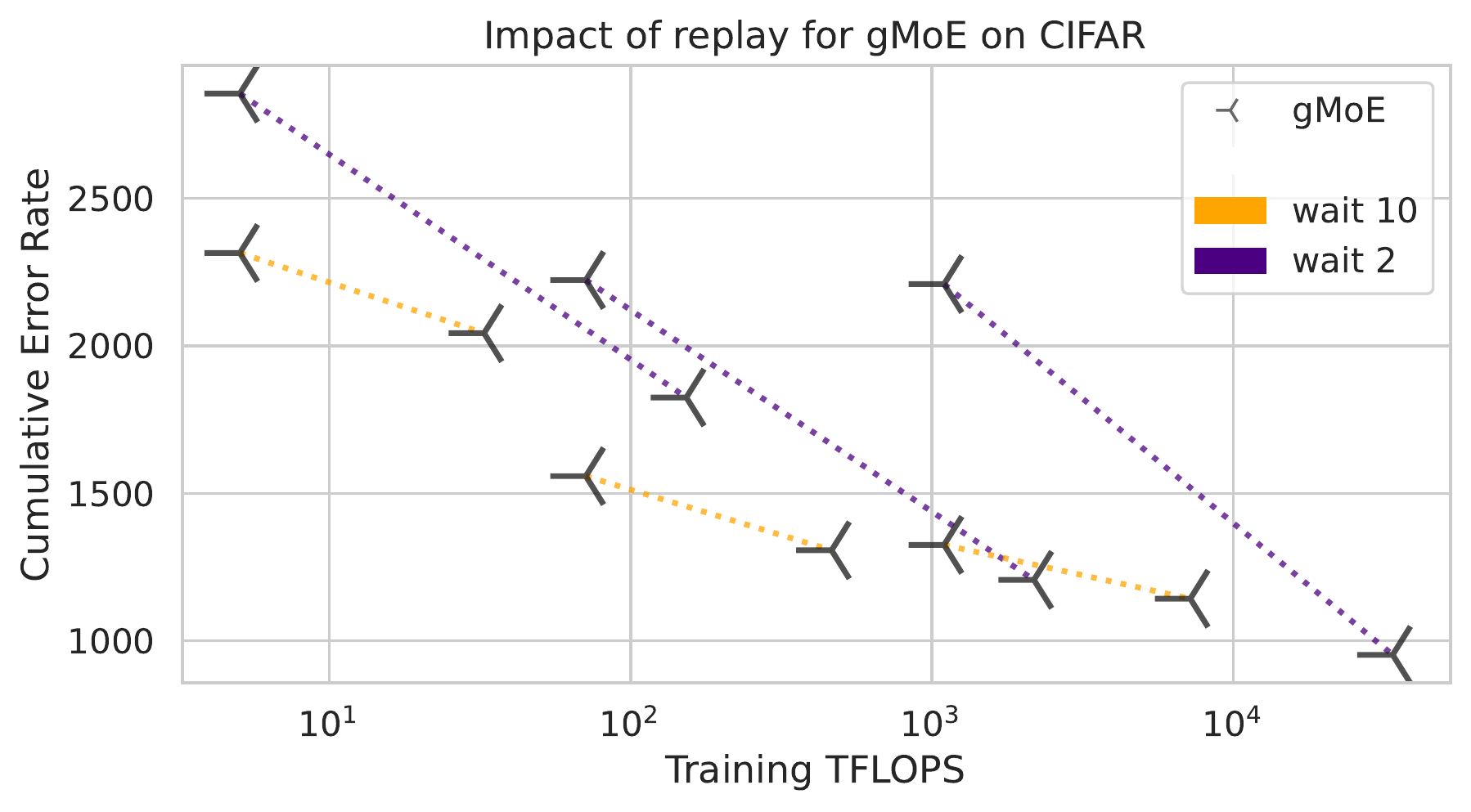}
\includegraphics[width=.45\textwidth]{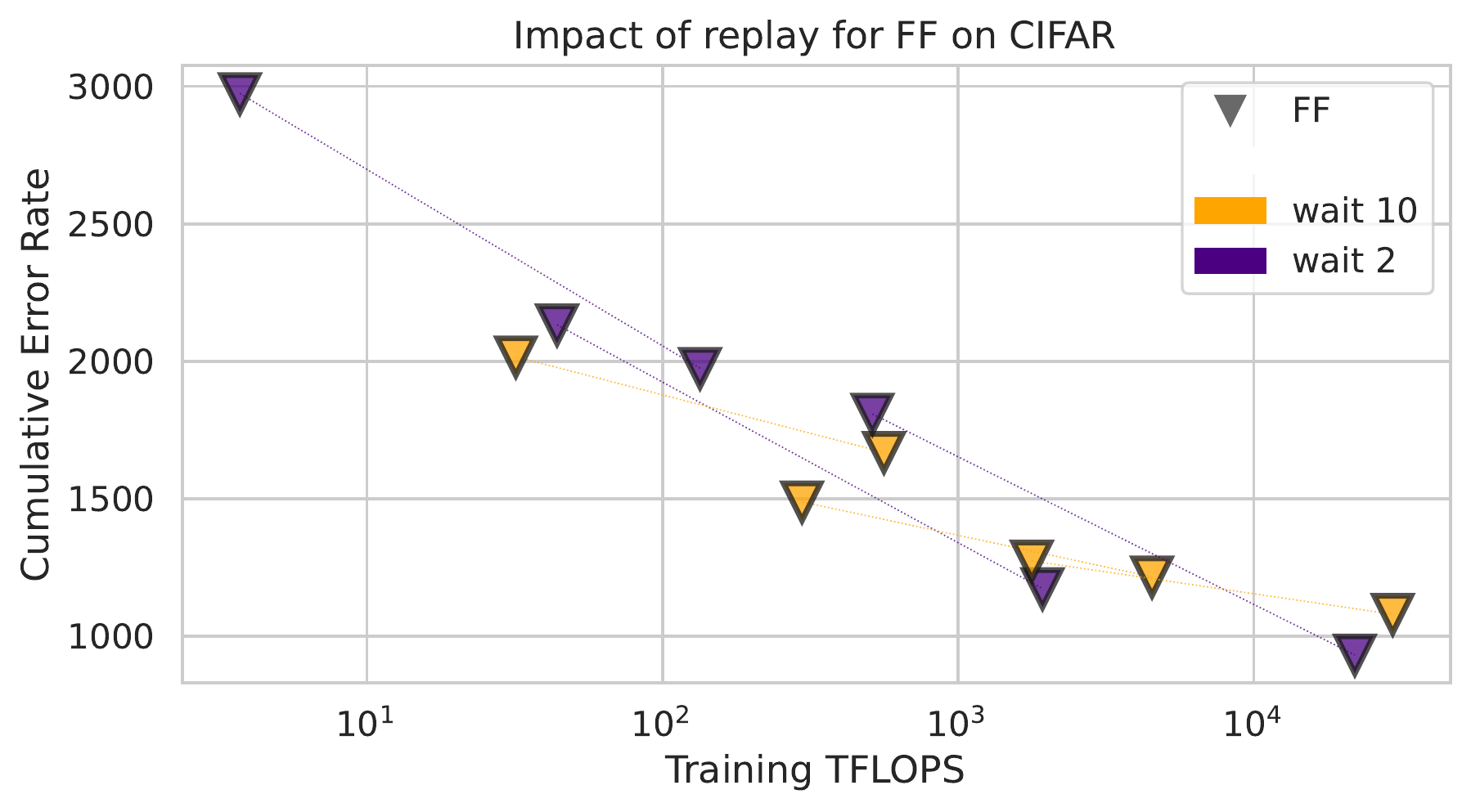}

  \vspace{-10pt}
  \caption{\small Impact of Replay across different methods and waiting times for CIFAR-10.}
  \label{fig:cifar_replay_appendix}
\end{figure*}

\newpage
\section{Full LM Results} \label{app:lm}

Below we present the full quantitative results for our language modeling experiments. 

\begin{table}[h!]
\fontsize{9}{9}\selectfont 
\renewcommand{\arraystretch}{1.5}

\centering
\begin{tabular}{cccccccccccc}
\toprule
& & \multicolumn{5}{c}{\emph{\textcolor{blue}{Base} model perplexity}} & \multicolumn{5}{c}{\emph{\textcolor{orange}{Large} model perplexity}} \\
setting & \# experts & $|\theta|$ & $t_0$ & $t_1$ & $t_2$ & $t_3$ & $|\theta|$ & $t_0$ & $t_1$ & $t_2$ & $t_3$ \\
\midrule
\Single\_w1 & 4 & 65M & 28.57 & 27.45 & 26.91 & 26.53 & 210M & 22.47 & 21.62 & 20.84 & 20.54 \\
\Single\_w1 & 8 & 91M & 26.29 & 25.29 & 24.74 & 24.40 & 323M & 21.52 & 20.38 & 19.64 & 19.22 \\
\Single\_w1 & 12 & 116M & 25.63 & 24.70 & 24.17 & 23.78 & 436M & 21.63 & 20.26 & 19.44 & 18.98 \\
\Single\_w3 & 8 & 91M & * & * & 25.18 & & 323M & * & * & 19.29 &  \\
\Single\_w3 (3x steps) & 8 & 91M & * & * & 24.21 & & 323M & * & * & 18.48 &  \\
\Single\_w4 & 12 & 116M & * & * & * & 24.41 & 436M & * & * & * & 19.01 \\
\Single\_w4 (4x steps) & 12 & 116M & * & * & * & 22.87 & 436M & * & * & * & 17.70 \\
\cdashlinelr{1-12}
\multirow{2}{*}{\Ens\_w1}  & 4@2 & 130M & 26.20 & 25.12 & 24.57 & 24.35 & 420M & 20.32 & 19.55 & 19.14 & 18.92 \\
 & 4@4 & 260M & 25.03 & 24.03 & 23.45 & 23.29 & 840M & 19.27 & 18.52 & 18.22 & 18.07 \\
\Ens\_w3  & 4@2 & 130M & * & * & 25.52 &  & 420M & * & * & 19.11 &  \\
\Ens\_w3 (3x steps)  & 4@2 & 130M & * & * & 24.30 &  & 420M & * & * & 18.04 &  \\
\Ens\_w4  & 4@2 & 130M & * & * & * & 25.49 & 420M & * & * & * & 19.03 \\
\Ens\_w4 (4x steps)  & 4@2 & 130M & * & * & * & 24.13 & 420M & * & * & * & 17.84 \\
\cdashlinelr{1-12}
\multirow{4}{*}{\gEns\_w1} & 4@1 & 65M & 28.57 & & & & 210M & 22.47 & & & \\
& 4@2 & 130M & & 26.27 & & & 420M & & 20.25 & & \\
& 4@3 & 195M & & & 25.41 & & 630M & & & 19.49 & \\
& 4@4 & 260M & & & & 25.01 & 840M & & & & 19.18 \\
\cdashlinelr{1-12}
\multirow{4}{*}{\gMoE\_w1} & 4 & 65M & 28.57 & & & & 210M & 22.47 & & & \\
& 6 & 78M & & 26.46 & & & 266M & & 21.22 & & \\
& 8 & 91M & & & 25.66 & & 323M & & & 20.39 & \\
& 12 & 116M & & & & 25.28 & 436M & & & & 20.15 \\
\bottomrule
\end{tabular}
\vspace{2pt}
\caption{Large scale language modeling results. For \Ens{} and \gEns, 4@3 means 3 components in the ensemble, each of which has 4 experts per block, for instance. }
\label{tab:lm}
\vspace{-10pt}
\end{table}

\newpage

\section{Extended Figure 1}
In this section, we add runs with replay to Fig. \ref{fig:alma_context}. We note that runs with replay are not directly comparable to runs without replay, because they have a higher computational cost. Indeed, runs that fine-tune every 10 chunks cost \textbf{4.5x} the cost of non-replay runs, and runs fine-tuning every chunk cost \textbf{122.5x}. 

\begin{figure}[h!]

  \centering
\includegraphics[width=.75\textwidth]{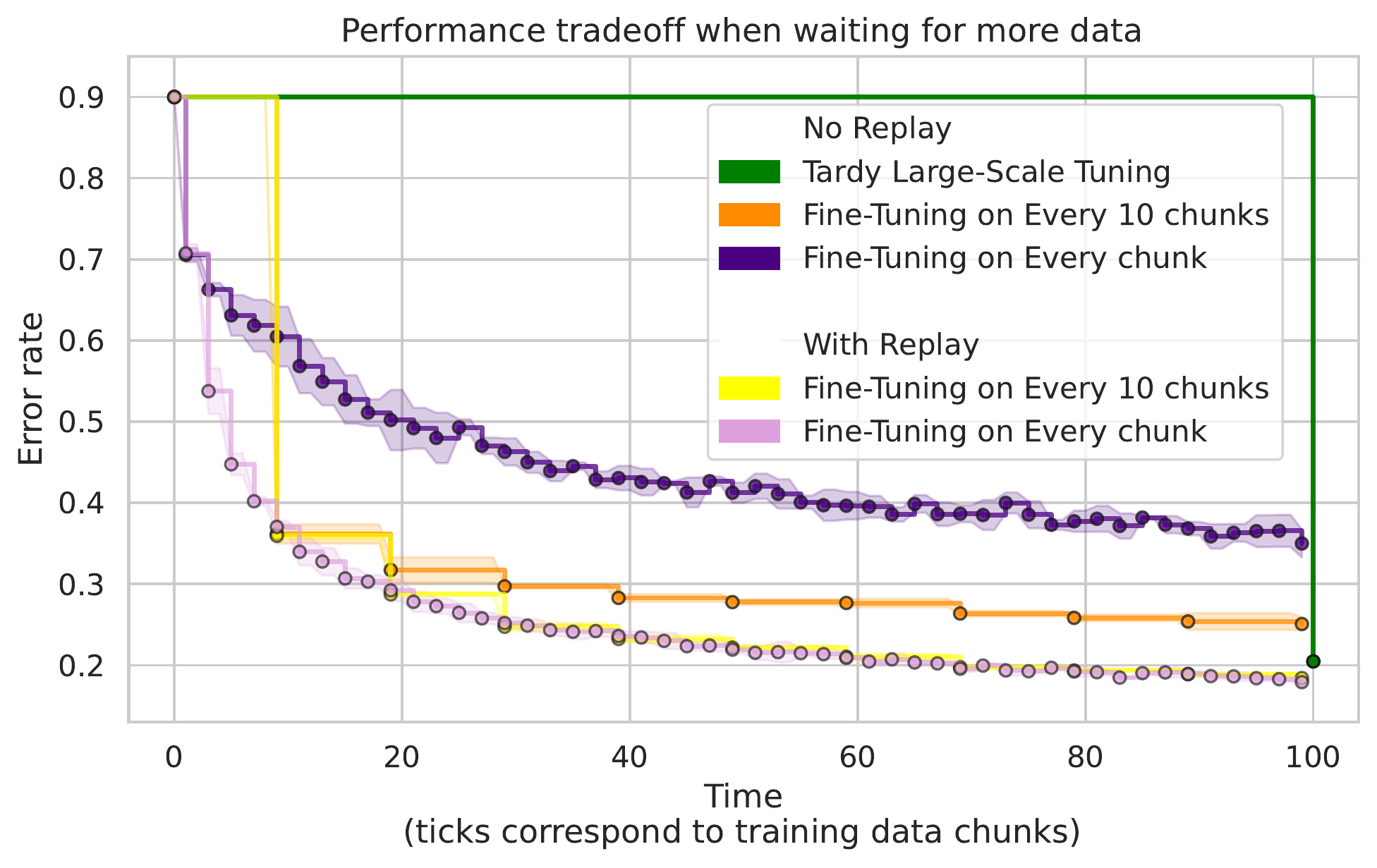} \\
\includegraphics[width=.75\textwidth]{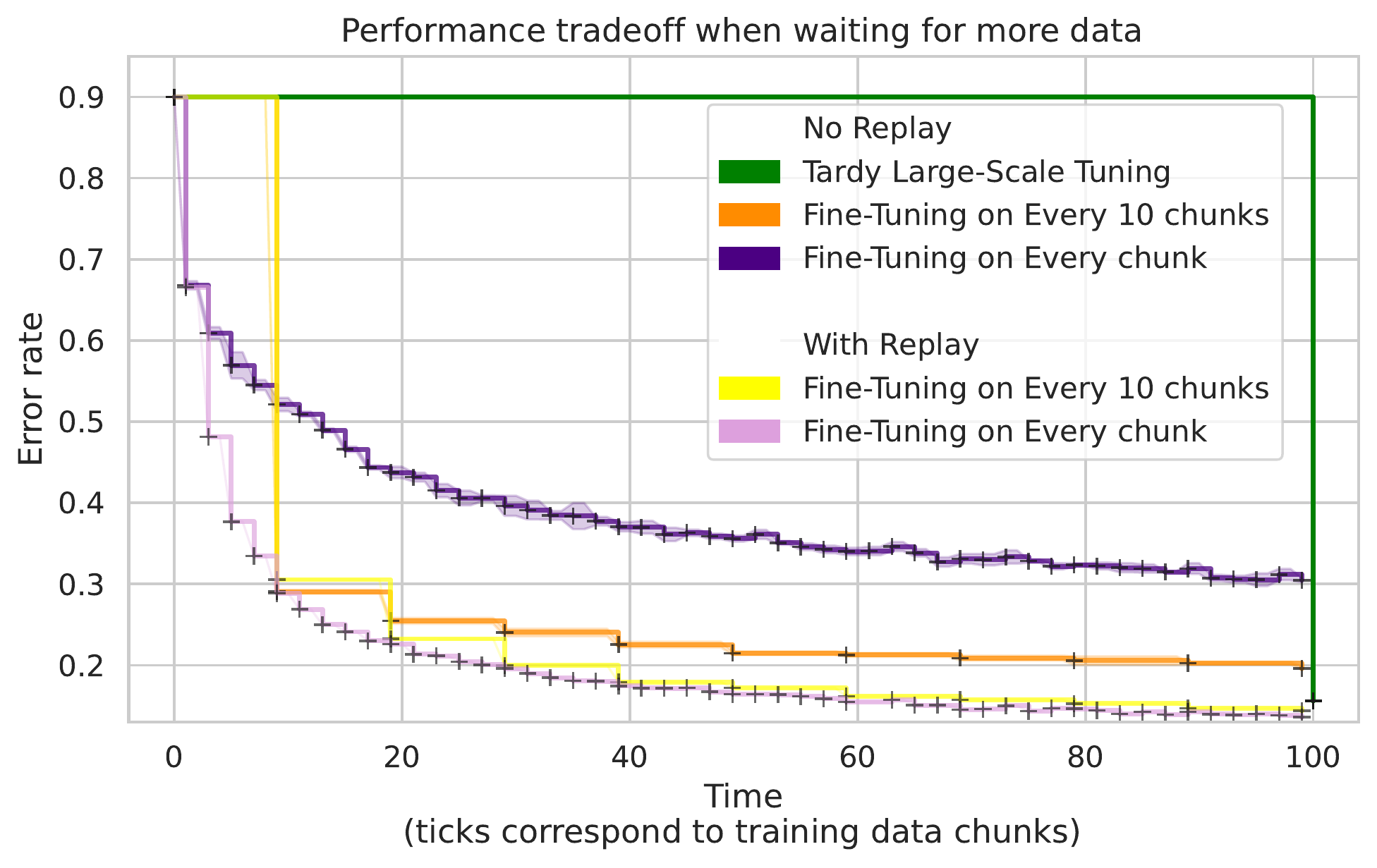}
  \caption{\small Fixed architecture runs (top) and growing ensemble runs (bottom) }  \label{fig:figure_1_more}
\end{figure}

\end{document}